\documentclass{article}

\usepackage{microtype}
\usepackage{graphicx}
\usepackage{subfigure}
\usepackage{booktabs} 

\usepackage{hyperref}

\PassOptionsToPackage{noend}{algorithmic} 

\usepackage[accepted]{for_arxiv}

\icmltitlerunning{Non-approximate Inference for Collective Graphical Models on Path Graphs via Discrete Difference of Convex Algorithm}

\usepackage{amssymb, bm, ascmac, amsthm, multirow, bigdelim, comment, mathtools}
\usepackage{algorithmic}
\allowdisplaybreaks[1]

\newcommand{\flowG}{\mathcal{G}}
\newcommand{\flowV}{\mathcal{V}}
\newcommand{\flowE}{\mathcal{E}}

\newcommand{\bn}{\bm{n}}

\newcommand{\bx}{\bm{x}}
\newcommand{\by}{\bm{y}}
\newcommand{\bz}{\bm{z}}

\newcommand{\bX}{\bm{X}}

\newcommand{\bphi}{\bm{\phi}}

\newcommand{\dI}{\mathbb{I}}
\newcommand{\dL}{\mathbb{L}}
\newcommand{\dZ}{\mathbb{Z}}
\newcommand{\dZp}{\mathbb{Z}_{\geq 0}}

\newcommand{\defeq}{\coloneqq}

\newcommand{\tbl}{n}
\newcommand{\btbl}{\bm{\tbl}}

\newcommand{\bheta}{\bm{\eta}}

\newtheorem{theorem}{Theorem}
\newtheorem*{theorem*}{Theorem}
\newtheorem{proposition}{Proposition}
\newtheorem*{proposition*}{Proposition}
\newtheorem{definition}[theorem]{Definition}
\newtheorem*{definition*}{Definition}

\newtheorem*{corollary*}{Corollary}

\newtheorem*{lemma*}{Lemma}
\newtheorem{assumption}{Assumption}
\newtheorem*{assumption*}{Assumption}

\renewcommand{\algorithmicrequire}{\textbf{Input:}}
\renewcommand{\algorithmicensure}{\textbf{Output:}}

\usepackage{cleveref}
\usepackage{autonum}
\DeclareMathOperator*{\tsum}{\textstyle\sum}

\crefname{equation}{}{}
\crefname{figure}{Figure}{Figures}
\DeclareMathOperator{\dist}{dist}
\DeclareMathOperator{\targmin}{arg~min}

\begin{document}

\twocolumn[
\icmltitle{Non-approximate Inference for Collective Graphical Models on Path Graphs \\ via Discrete Difference of Convex Algorithm}




\begin{icmlauthorlist}
\icmlauthor{Yasunori Akagi}{ev}
\icmlauthor{Naoki Marumo}{cs}
\icmlauthor{Hideaki Kim}{ev}
\icmlauthor{Takeshi Kurashima}{ev}
\icmlauthor{Hiroyuki Toda}{ev}
\end{icmlauthorlist}

\icmlaffiliation{ev}{NTT Service Evolution Laboratories}
\icmlaffiliation{cs}{NTT Communication Science Laboratories}

\icmlcorrespondingauthor{Yasunori Akagi}{yasunori.akagi.cu@hco.ntt.co.jp}

\icmlkeywords{Machine Learning, ICML}

\vskip 0.3in
]



\printAffiliationsAndNotice{}  

\begin{abstract}
  The importance of aggregated count data, which is calculated from the data of multiple individuals, continues to increase. Collective Graphical Model (CGM) is a probabilistic approach to the analysis of aggregated data. One of the most important operations in CGM is maximum a posteriori (MAP) inference of unobserved variables under given observations. Because the MAP inference problem for general CGMs has been shown to be NP-hard, an approach that solves an approximate problem has been proposed. However, this approach has two major drawbacks. First, the quality of the solution deteriorates when the values in the count tables are small, because the approximation becomes inaccurate. Second, since continuous relaxation is applied, the integrality constraints of the output are violated. To resolve these problems, this paper proposes a new method for MAP inference for CGMs on path graphs. First we show that the MAP inference problem can be formulated as a (non-linear) minimum cost flow problem. Then, we apply Difference of Convex Algorithm (DCA), which is a general methodology to minimize a function represented as the sum of a convex function and a concave function. In our algorithm, important subroutines in DCA can be efficiently calculated by minimum convex cost flow algorithms. Experiments show that the proposed method outputs higher quality solutions than the conventional approach. 
\end{abstract}

\section{Introduction}
In recent years, the importance of aggregated count data, which is calculated from the data of multiple individuals, has been increasing \cite{Tanaka2019, Zhang2020}.
Although technologies for acquiring individual data such as sensors and GPS have greatly advanced, it is still very difficult to handle individual data due to privacy concerns and the difficulty of tracking individuals.
However, there are many situations where data aggregated from multiple individuals can be obtained and utilized easily.
For example, Mobile Spatial Statistics \cite{Terada2013}, which is the hourly population data of fixed-size square grids calculated from cell phone network data in Japan, are available for purchase; such data is being used for disaster prevention and urban planning \cite{Suzuki2013}. 
In traffic networks, traffic volume data at each point can be obtained more easily by sensors or cameras than the trajectories of individual cars, and the data is useful for managing traffic congestion \cite{Morimura2013,Zhang2017}. 

Collective Graphical Model (CGM) \cite{Sheldon2011} is a probabilistic model to describe aggregated statistics of multiple samples drawn from a graphical model. 
CGM makes it possible to conduct various practical tasks on aggregate count data, such as estimating movements from population snapshots, parameter learning, interpolation and denoising of count tables. 
In this paper, we focus on the case where the underlying graphical model is on a path graph.
CGMs on path graphs are particularly important because they treat time series data in which the states of interest follow Markov chains. 
In fact, most of the real-world applications of CGMs utilize CGMs on path graphs to represent the collective movement of humans and animals \cite{Du2014, Sun2015, Akagi2018}.

\begin{figure*}[t]
  \begin{center}
  \centerline{\includegraphics[width=1.9\columnwidth]{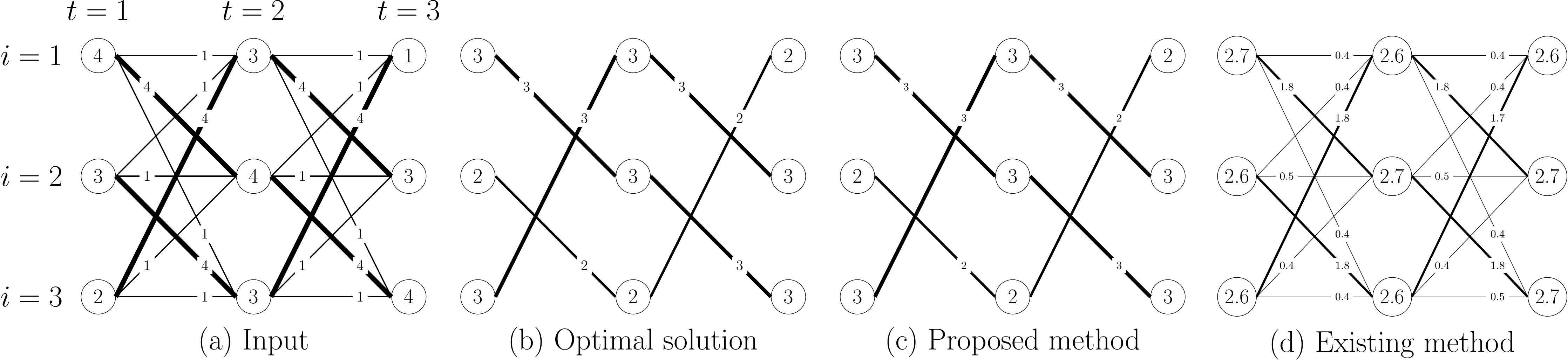}}
  \label{fig: output comparison}
  \caption{Comparison of MAP inference results obtained by the proposed method and the existing method. 
  (a) Input of the MAP inference problem. The values written in nodes are observations and the values written on edges are potentials. As the noise distribution, we use Gaussian distribution. 
  (b) An optimal solution obtained by brute force search. Note that brute force search is possible only when input size is extremely small. 
  (c) The output of the proposed method.
  (d) The output of the existing method, non-linear belief propagation \cite{Sun2015}. 
  The values written in nodes are node count table values and the values written on edges are edge count table values. 
  The total population is set to be 8 in all methods. As we can see, the proposed method outputs integer-valued and sparse optimal solutions while the existing method does not.}
  \end{center}
  \vskip -0.2in
\end{figure*}

One of the most important operations in CGM is maximum a posteriori (MAP) inference.  
MAP inference is the discrete (combinatorial) optimization problem of finding an assignment of unobserved variables that maximizes posterior probability under given observations.
MAP inference makes it possible to interpolate missing values of aggregated data and to estimate more detailed information that lies behind the observations.
Unfortunately, MAP inference for general CGMs has been shown to be NP-hard \cite{Sheldon2013a} and thus is difficult to solve exactly and efficiently.
Therefore, an alternative approach that solves an approximate problem, which is derived by applying Stirling's approximation and continuous relaxation, has been proposed \cite{Sheldon2013a}. 
Subsequent studies have focused on solving this approximate problem efficiently \cite{Sun2015, Vilnis2015,Nguyen2016, Singh2020}.

However, there are inherent problems with this approach of solving the approximate problem. 
First, this approach tends to output a solution with low posterior probability when the values in the count tables are small, because Stirling’s approximation, $\log x! \approx x \log x - x$, is inaccurate when $x$ is small.
This situation frequently occurs when the number of values that each variable in the graphical model takes is large, or when the total population is small.
Second, since continuous relaxation is applied, the integrality constraints of count table values are violated in the output. 
As a result, values that should be integers (e.g., the number of people) are no longer integers, which not only reduces interpretability, but also makes the output less sparse, resulting in high memory consumption to maintain the output.
It is possible to obtain integer-valued results by rounding output, but this rounding process destroys the consistency of the counts among the aggregated data. 
For example, the sum of the count table values at each node may not match the total population.

In this paper, we propose a new method for MAP inference for CGMs on path graphs to resolve these issues.
First, we show that the problem can be formulated as a combinatorial optimization problem called the (nonlinear) minimum cost flow problem (MCFP).
MCFP is known to be an optimization problem with good properties and can be solved efficiently when the cost function on each edge satisfies the  discrete convex conditions \cite{Ahuja1993}.
However, the MCFP in our settings is difficult to solve. 
because the cost functions of several edges do not satisfy the discrete convex condition. 
To deal with this, we utilize the Difference of Convex Algorithm (DCA) \cite{Le2018}.
DCA is a framework to minimize a function expressed as the sum of a convex function and a concave function. 
A solution is obtained by repeatedly minimizing a surrogate function that upper-bounds the objective function.
We show that in our problem, the surrogate function can be easily computed in closed form, and that the surrogate function can be minimized efficiently by minimum convex cost flow algorithms. 
The proposed algorithm decreases the objective function value monotonically in each iteration and terminates in a finite number of iterations.

The proposed method has several practical advantages.
First, since the proposed method does not use Stirling's approximation, it offers accurate inference even when the values in the count tables are small. 
This makes it possible to output solutions with much higher posterior probability than the approximation-based approach.  
Second, because the proposed method does not apply continuous relaxation, the obtained solution is guaranteed to be integer-valued, which results in sparse and interpretable outputs. 
Figure \ref{fig: output comparison} illustrates the difference in the output of the proposed and existing methods. As shown, the output of the proposed method is integer-valued and sparse, while that of the existing method is not. 
In Section \ref{sec: experiments}, we show results gained from synthetic and real-world datasets; they indicate that the proposed method outputs higher quality solutions than the existing approach. 
We show that the superiority of the proposed method is much greater when the overall population is not very large or the number of states on nodes in the graphical model is large.


\section{Preliminaries}
\subsection{Collective Graphical Model (CGM)}
Collective Graphical Model (CGM) is a probabilistic generative model that describes the distributions of aggregated statistics of multiple samples drawn from a certain graphical model \cite{Sheldon2011}. 
Let $G=(V, E)$ be an undirected tree graph (i.e., a connected graph with no cycles) with $V = [N] \defeq \{1, 2, \ldots, N\}$. 
We consider a pairwise graphical model over discrete random variable $\bX \defeq (X_1, \ldots, X_{N})$ defined by
\begin{align}
  \Pr(\bX=\bx) = \frac{1}{Z} \prod_{(u, v) \in E} \phi_{u, v}(x_u, x_v), 
\end{align}
where $\phi_{uv}(x_u, x_v)$ is a local potential function on edge $(u, v)$ and $Z \defeq \sum_{\bx} \prod_{(u, v) \in E} \phi_{u, v}(x_u, x_v)$ is a partition function for normalization. 
In this paper, we assume that $x_u$ takes values on the set $[R]$ for all $u \in V$. 

We draw ordered samples $\bX^{(1)}, \ldots, \bX^{(M)}$ independently from the graphical model. $M$ is called the total population. 
We define \emph{node contingency table} (count table) $\btbl_u \defeq (\tbl_u(i) \mid i \in [R])$ for $u \in V$ and \emph{edge contingency table} $\btbl_{u, v} \defeq (\tbl_{u, v}(i, j) \mid i, j \in [R])$ for $(u, v) \in E$, which are the vectors whose entries are the number of occurrences of particular variable settings: 
\begin{align}
  \tbl_u(i) &\defeq \sum_{m=1}^M \mathbb{I} (X_u^{(m)}=i), \\
  \tbl_{u, v}(i, j) &\defeq \sum_{m=1}^M \mathbb{I} (X_u^{(m)}=i,\ X_v^{(m)}=j), 
\end{align}
where $\mathbb{I}(\cdot)$ is the indicator function. 
In CGM, observations 
$\by \defeq \left( (\by_u)_{u \in V}, (\by_{u, v})_{(u, v) \in E} \right)$ 
are generated by adding noise to contingency tables $\btbl :=\left( (\btbl_u)_{u \in V}, (\btbl_{u, v})_{(u, v) \in E} \right)$. 

The MAP inference problem for CGM is to find $\btbl^*$ which maximizes the posterior probability $\Pr(\btbl|\by)$. 
Since $\Pr(\btbl|\by) = \Pr(\btbl, \by)/\Pr(\by)$ from Bayes' rule, it suffices to maximize the joint probability $\Pr(\btbl, \by) = \Pr(\btbl) \cdot \Pr(\by | \btbl)$, where $\Pr(\by | \btbl)$ is the noise distribution associated with observation. $\Pr(\btbl)$ is called CGM distribution and calculated as follows \cite{Sun2015}:
{
\begin{align}
  \Pr(\btbl) ={}& F(\btbl) \cdot \dI (\btbl \in \dL_M^{\dZ}), \label{formula: Pr}\\
  F(\btbl) \defeq{}& \frac{M!}{Z^M} \cdot \frac{\prod_{u \in V} \prod_{i \in [R]} \left(\tbl_u(i)! \right)^{\nu_u - 1}}{\prod_{(u, v) \in E} \prod_{i, j \in [R]} \tbl_{u, v}(i, j)!} \\ 
  &\quad \cdot \prod_{(u, v) \in E} \prod_{i, j \in [R]} \phi_{u, v}(i, j)^{\tbl_{u, v}(i, j)}, \label{formula: F}\\
  \dL_M^\dZ \defeq{}&
  \Bigl\{ \btbl \in \dZp^{|V|R + |E|R^2} \ \Bigl|\  M = \! \tsum_{i \in [R]} \! \tbl_u(i) \  (\forall u \in V),\\[-0.2\baselineskip]
  & \tbl_u(i) = \! \tsum_{j \in [R]} \! \tbl_{u, v}(i, j) \ (\forall (u, v) \in E,\  i \in [R]) \Bigr\}. \label{formula: L}
\end{align}}%
Here, $\dL_M^\dZ$ is the set of contingency tables $\btbl$ that satisfy the consistency of counts among the number of samples $M$, the node contingency tables $\btbl_u$, and the edge contingency tables $\btbl_{u, v}$. 
$\nu_u$ is the degree of node $u$ in $G$. 
Using the above notations, the MAP inference problem can be written as 
\begin{align}
\min_{\btbl \in \dL_M^\dZ} \   -\log F(\btbl) - \log\Pr(\by|\btbl). \label{formula: MAP}
\end{align}

\subsection{CGMs on Path Graphs}
We explain the details of CGMs on path graphs, which is the main topic of this paper. 
Path graph $P_n$ is an undirected graph whose vertex set is $V = [N]$ and edge set is $E =\{ (t, {t+1}) \mid t \in [N-1] \}$. 
A graphical model (not CGM) on path graph is the most basic graphical model that represents a time series generated by a Markov model; that is, the current state depends only on the previous state. 
A CGM on a path graph represents the distribution of aggregated statistics when there are many individuals whose state transition is determined by a Markov model. 
In the rest of this paper, we use the notation $ \tbl_{ti} \defeq \tbl_t(i)$, $\tbl_{t i j} \defeq \tbl_{t, t+1} (i, j)$, and $\phi_{tij} \defeq \phi_{t, t+1} (i, j)$ for simplicity. 

\begin{figure}[t]
  \begin{center}
  \centerline{\includegraphics[width=0.8\columnwidth]{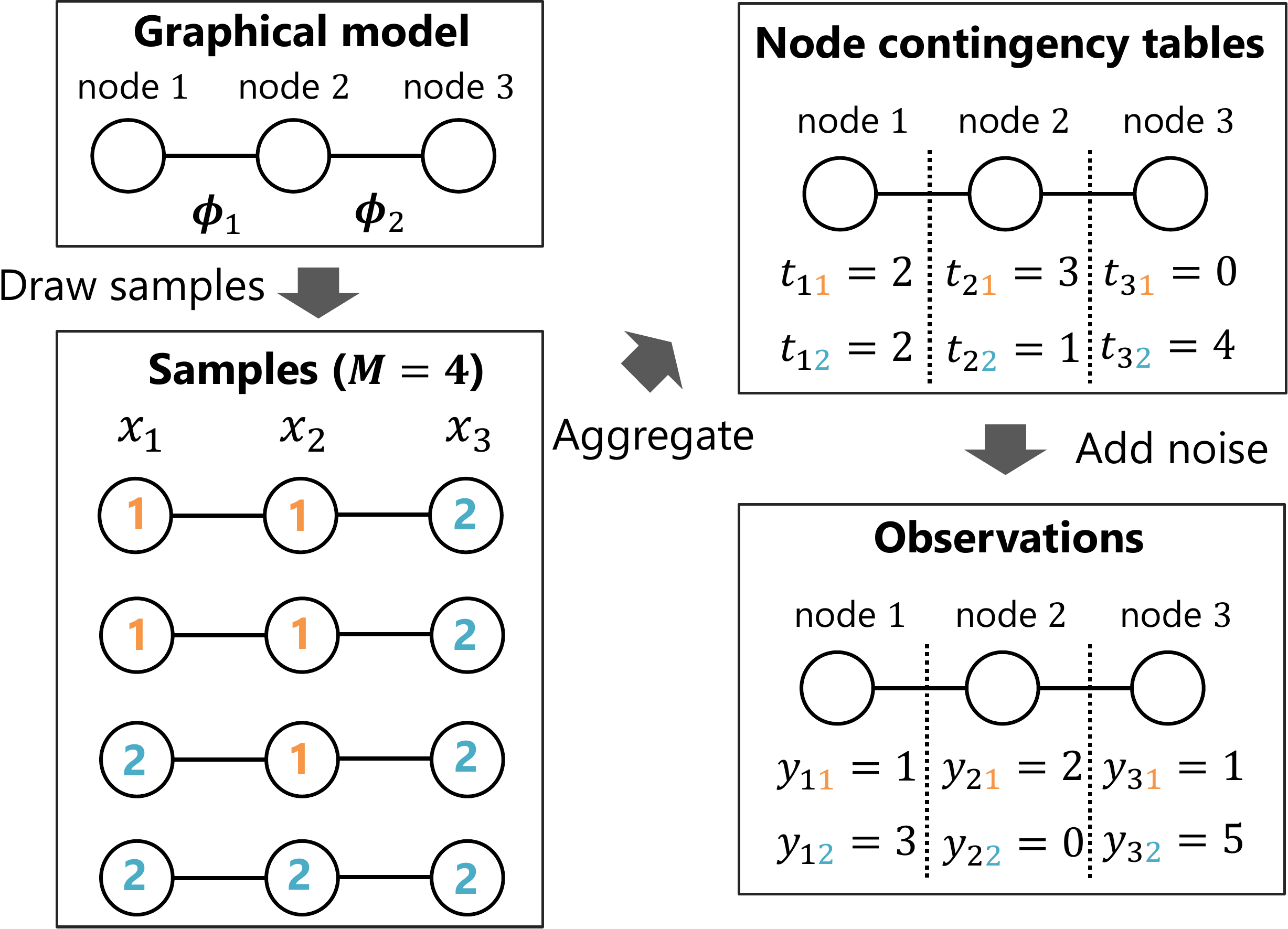}}
  \caption{An example of generation process of observations in CGM on a path graph when $N=3, R=2, M=4$. }
  \label{fig: generation}
  \end{center}
  \vskip -0.2in
\end{figure}

To develop the concrete form of the MAP inference problem, we need to decide the observation model $\Pr(\by|\bn)$.
In this paper, we consider the \emph{node observation model}, a model in which observations are obtained at each node with independent noise. In this case, observation $\by$ consists of $y_{ti}\ (t \in [N],\ y \in [R])$, and its distribution is given by
$
\Pr(\by | \btbl ) = \prod_{t \in [N]} \prod_{i \in [R]} p_{ti}(y_{ti}|\tbl_{ti})
$, where $p_{ti}$ is the noise distribution.  
An additional assumption is described below. 
\begin{assumption} \label{assumption: log concave}
For $t \in [N]$ and $i \in [R]$, $\log p_{ti}(y|\tbl)$ is a concave function in $\tbl$. 
\end{assumption}
Assumption \ref{assumption: log concave} is a quite common assumption in CGM studies \cite{Sheldon2013a, Sun2015}.  Commonly used noise distributions such as Gaussian distribution 
$p_{ij}(y_{ij}|t_{ij}) = \frac{1}{\sqrt{2 \pi \sigma^2}}\exp \big( \frac{-(y_{ij} - t_{ij})^2}{2\sigma^2}\big)$ 
and Poisson distribution
$p_{ij}(y_{ij}|t_{ij}) = \frac{t_{ij}^{y_{ij}} \cdot \exp \left( -t_{ij} \right)}{y_{ij}!}$
satisfy Assumption \ref{assumption: log concave}. 
See Figure \ref{fig: generation}  for the relationship between these symbols.

We give an example drawn from human mobility. 
Consider a space that is divided into $R$ distinct areas, and $M$ people are moving around in the space. 
The time series of areas to which person $m$ belongs at each time step,  $\bX^{(m)} = \big( X^{(m)}_1, \ldots, X^{(m)}_N \big)$, is determined by the graphical model 
$p(\bx) = \frac{1}{Z} \prod_{t=1}^{N-1} \phi_{t x_t x_{t+1}}$. 
Here, 
$\phi_{t x_t x_{t+1}}$
is the affinity between two areas $x_t$ and $x_{t+1}$ at time step $t \rightarrow t+1$.
$\tbl_{ti}$ represents the number of people in area $i$ at time step $t$, and
$\tbl_{tij}$ represents the number of people who moved from area $i$ to area $j$ at time step $t \rightarrow t+1$.
We have noisy observations $y_{ti}$ for $t \in [N]$ and $i \in [R]$, which are generated by adding noise to $\tbl_{ti}$. 
The MAP inference problem we want to solve is to find the true population of each area at each time step, 
$\{\tbl_{ti}\}_{t \in [N], i \in [R]}$, and the true number of people moving between areas, 
$\{\tbl_{tij}\}_{t \in [N-1], i, j \in [R]}$,
with the highest posterior probability given the observation $\{y_{ti}\}_{t \in [N], i \in [R]}$.

From \cref{formula: Pr,formula: F,formula: L,formula: MAP}, the MAP inference problem for CGMs on path graphs can be written as follows:
\begin{align} \label{problem: MAP on path}
  \begin{aligned}
  \min_{\btbl}& && \sum_{t=1}^{N-1} \sum_{i, j \in [R]} f_{tij}(\tbl_{tij})
  +\sum_{t=2}^{N-1} \sum_{i \in [R]} g(\tbl_{ti}) \\
  &&&
  + \sum_{t=1}^{N} \sum_{i \in [R]} h_{ti}(\tbl_{ti}), \\
  \mathrm{s.t.}&&& \sum_{i \in [R]} \tbl_{ti} = M \quad (t \in [N]), \\
  &&& \sum_{j \in [R]} \tbl_{tij} = \tbl_{ti} \quad (t \in [N-1],\ i \in [R]), \\
  &&& \sum_{i \in [R]} \tbl_{tij} = \tbl_{i+1, j} \quad (t \in [N-1],\ j \in [R]), \\
  &&&  \tbl_{tij}, \tbl_{ti} \in \dZp, 
  \end{aligned}
\end{align}
where 
\begin{align}
  f_{tij}(z) &\defeq \log z! - z \cdot \log \phi_{tij}, \quad
  g(z) \defeq - \log z!, \\
  h_{ti}(z) &\defeq - \log \left[ p_{ti}(y_{ti}|z) \right]. 
\end{align}
For the details of the derivation, please see Appendix. 
Let $\mathcal{P}(\btbl)$  denote the objective function of problem ($\ref{problem: MAP on path}$). 

\section{Proposed Method} \label{sec: Proposed Method}

\subsection{Formulation as Minimum Cost Flow Problem} \label{subsec: formulation}
First, we show that problem (\ref{problem: MAP on path}) can be formulated as the (non-linear) minimum cost flow problem (MCFP). 
MCFP is a combinatorial optimization problem on the directed graph $\flowG=(\flowV, \flowE)$. 
Each node $i \in \flowV$ has a supply value $b_{i} \in \mathbb{Z}$, and
each edge $(i, j) \in \flowE$ has a cost function $c_{ij}: \mathbb{Z}_{\geq 0} \to \mathbb{R} \cup \{ + \infty\}$.
MCFP is the problem of finding a minimum cost flow on $\flowG$ that satisfies the supply constraints at all nodes. 
MCFP can be described as follows:
\begin{align}\label{problem: minimum_cost_flow}
\begin{aligned}
\min_{\bz \in \mathbb{Z}^{|\flowE|}}& && \sum_{(i, j) \in \flowE} c_{ij}(z_{ij}) \\
\mathrm{s.t.}& && \sum_{j:(i, j)\in \flowE} z_{ij} - \sum_{j:(j, i)\in \flowE} z_{ji} = b_{i} \quad (i \in \flowV). 
\end{aligned}
\end{align}
Note that $\bz$ takes only integer values (i.e., $\bz \in \mathbb{Z}^{|\flowE|}$).

To formulate the MAP inference problem (\ref{problem: MAP on path}) as MCFP, we construct an MCFP instance by Algorithm \ref{alg: graph construction}. 
\begin{proposition} \label{prop: MCFP}
  For $\btbl^*$ defined by $\tbl^*_{ti} \defeq z^*_{u_{t, i} w_{t,i}}$ and $\tbl^*_{tij} \defeq z^*_{w_{t, i} u_{t+1, j}}$, where $\bz^*$ is an optimal solution of the MCFP instance constructed by Algorithm \ref{alg: graph construction}, $\btbl^*$ is an optimal solution of problem (\ref{problem: MAP on path}).
\end{proposition}
Our proof is given in the Appendix. 
Figure \ref{fig: flow formulation} illustrates an example of an MCFP instance constructed by Algorithm \ref{alg: graph construction} where $N=3$ and $R=2$. 
This MCFP can be interpreted as the problem of finding a way to push $M$ flows from node $o$ to node $d$ with minimum cost.  


\begin{algorithm}[t]
  \caption{The MCFP instance construction algorithm} \label{alg: graph construction}
  \begin{algorithmic}[1]
  \STATE{$\flowV \defeq \{ o, d \} \cup (\cup_{t \in [N]}  ( \mathcal{U}_{t} \cup  \mathcal{W}_{t} ) ) $, 
  where $\mathcal{U}_{t} \defeq \{ u_{t, i} \}_{i \in [R]}$, $\mathcal{W}_{t} \defeq \{ w_{t, i} \}_{i \in [R]}$. }
  \STATE{Add edge $(o, u_{1, i}, 0)$ for $i \in [R]$. \\ // $(u, v, c(z))$ represents an directed edge from node $u$ to node $v$ with cost function $c(z)$. } 
  \STATE{Add edge $(w_{N, i}, d, 0)$ for $i \in [R]$. }
  \STATE{Add edge $(u_{t, i}, w_{t, i}, h_{ti}(z))$ for $t \in \{ 1, N \}$, $i \in [R] $. }
  \STATE{Add edge $(u_{t, i}, w_{t, i}, g(z) + h_{ti}(z))$ for $t=2, \ldots, N-1$, $i \in [R]$. }
  \STATE{Add edge $(w_{t, i}, u_{t+1, i}, f_{tij}(z))$ for $t \in [N-1]$, $i, j \in [R]$. }
  \STATE{Set $b_o = M$, $b_d = -M$, $b_v = 0$ for $v \in V \setminus \{o, d\}$.}
   \end{algorithmic}
 \end{algorithm}

\begin{figure}[t]
  \begin{center}
  \centerline{\includegraphics[width=0.9\columnwidth]{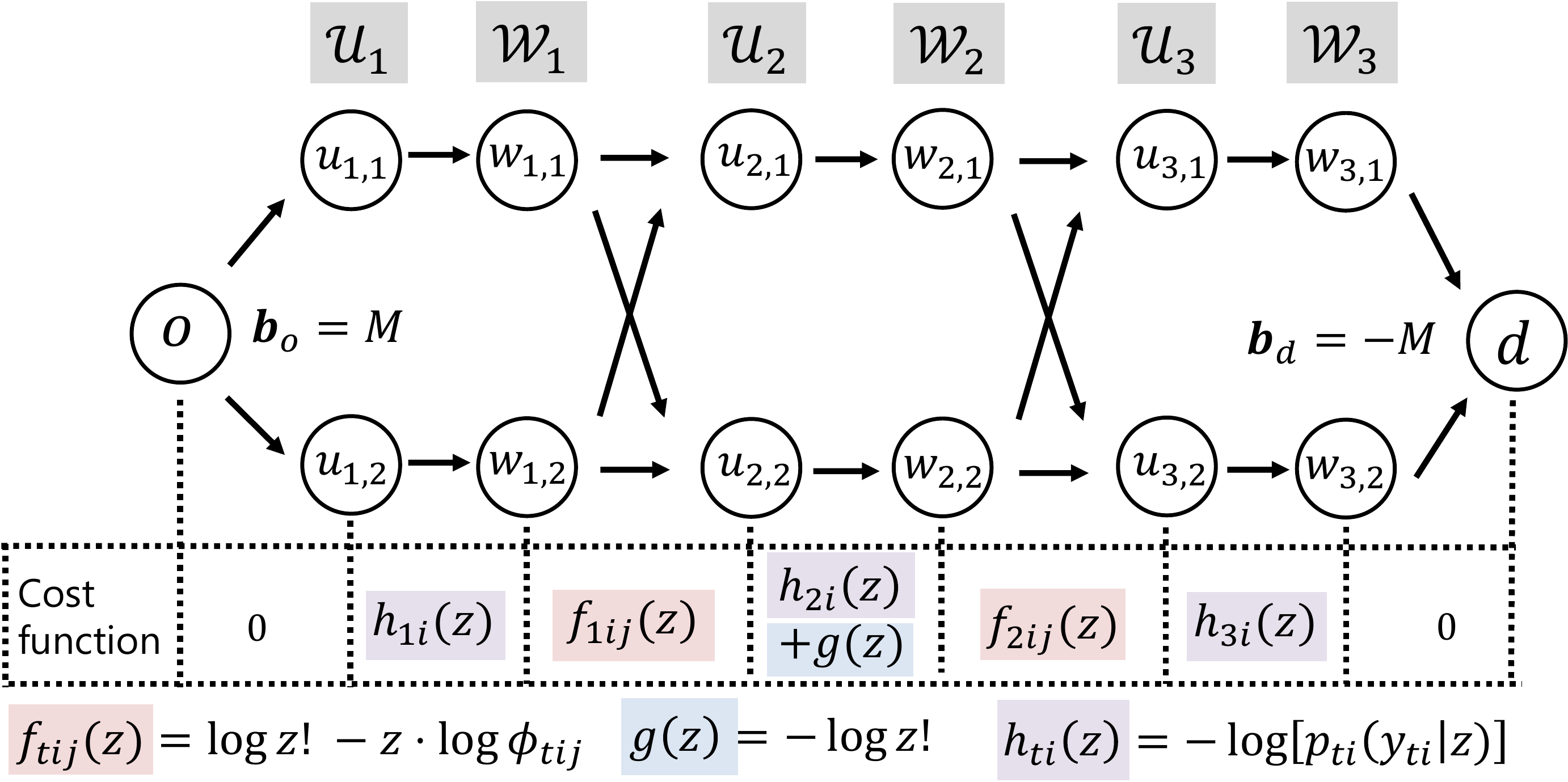}}
  \caption{An MCFP instance constructed by Algorithm \ref{alg: graph construction} when $N = 3$ and $R =2$. }
  \label{fig: flow formulation}
  \end{center}
  \vskip -0.2in
\end{figure}


\subsection{Discrete convexity of cost functions}
Proposition \ref{prop: MCFP} implies that we can obtain the optimal solution of the MAP inference problem (\ref{problem: MAP on path}) by solving the MCFP constructed above.
This, however, is not an easy task. 

In MCFP, the convexity of cost functions generally has a crucial impact on the computation complexity. 
\begin{definition}
    A function $f: \dZp \to \mathbb{R} \cup \{ +\infty \}$ is called a discrete convex function when $f(z+2) + f(z) \geq 2 \cdot f(z+1)$ for all $z \in \dZp$. 
    If $-f$ is a discrete convex function, $f$ is called a discrete concave function. 
\end{definition}
A subclass of MCFP in which all cost functions are discrete convex functions is called the Minimum Convex Cost Flow Problem (C-MCFP); it is  known to be efficiently solvable \cite{Ahuja1993}. 
However, the constructed MCFP is not an instance of C-MCFP. 
\begin{proposition} \label{prop: convexity}
  $f_{tij}$ is a discrete convex function. 
  Under Assumption \ref{assumption: log concave}, $h_{ti}$ is a discrete convex function. 
  $g$ is a discrete concave function. 
\end{proposition}
Proof is given in the Appendix. Thus, we cannot directly apply C-MCFP algorithms to our problem. 

\subsection{Application of DCA}
To overcome this issue, we utilize the idea of the Difference of Convex Algorithm (DCA). 
We first explain the core idea of DCA based on the description in \cite{Narasimhan2005}.
DCA is a general framework to solve the minimization problem $\min_{\btbl \in D} \mathcal{P}(\btbl) = \mathcal{Q}(\btbl) + \mathcal{R}(\btbl)$, where $\mathcal{Q}(\btbl)$ is a convex function and $\mathcal{R}(\btbl)$ is a concave function. 
DCA does this by using the following procedure to generate a feasible solution sequence $\btbl^{(1)}, \ldots, \btbl^{(s)}$ that satisfies $\mathcal{P}(\btbl^{(1)}) \geq \mathcal{P}(\btbl^{(2)}) \geq \cdots \geq \mathcal{P}(\btbl^{(s)})$. 
First, we set an arbitrary feasible solution in $D$ as $\btbl^{(1)}$. 
When we already have the sequence $\btbl^{(1)}, \ldots, \btbl^{(s)}$, we find a function $\bar{\mathcal{R}}^{(s)}(\btbl)$ that satisfies the following three conditions:
\begin{enumerate}
  \renewcommand{\labelenumi}{(\roman{enumi})}
  \item $\bar{\mathcal{R}}^{(s)}(\btbl^{(s)}) = \mathcal{R}(\btbl^{(s)})$, 
  \item $\bar{\mathcal{R}}^{(s)}(\btbl) \geq \mathcal{R}(\btbl) \ \ (\forall \btbl \in D)$, 
  \item $\bar{\mathcal{P}}^{(s)}(\btbl) \defeq \mathcal{Q}(\btbl) + \bar{\mathcal{R}}^{(s)}(\btbl)$ can be minimized efficiently in $D$. 
\end{enumerate}
Because $ \mathcal{R}(\btbl)$ is concave, by setting $\bar{\mathcal{R}}^{(s)}(\btbl) =  \mathcal{R}(\btbl^{(s)}) + \nabla{\mathcal{R}}(\btbl^{(s)}) \cdot (\btbl - \btbl^{(s)})$, which is a linear approximation of $ \mathcal{R}(\btbl)$ at $\btbl^{(s)}$, conditions (i)--(iii) hold. 
Using this function, we get a new feasible solution by $ \btbl^{(s+1)} = \targmin_{\btbl} \bar{\mathcal{P}}^{(s)}(\btbl) $. 
This can be done easily because condition (iii) holds. 
Then, because ${\mathcal{P}}(\btbl^{(s+1)}) \leq \bar{\mathcal{P}}^{(s)}(\btbl^{(s+1)}) \leq \bar{\mathcal{P}}^{(s)}(\btbl^{(s)}) = {\mathcal{P}}(\btbl^{(s)})$, we get $\mathcal{P}(\btbl^{(1)}) \geq \mathcal{P}(\btbl^{(2)}) \geq \cdots \geq \mathcal{P}(\btbl^{(s)}) \geq \mathcal{P}(\btbl^{(s+1)})$ by induction. 

We apply the framework of DCA to our MAP inference problem. 
We decompose $\mathcal{P}(\btbl)$, which is the objective function of (\ref{problem: MAP on path}), into $\mathcal{Q}(\btbl) = \sum_{t=1}^{N-1} \sum_{i,j \in [R]} f_{tij}(\tbl_{tij}) + \sum_{t=1}^{N} \sum_{i \in [R]} h_{ti}(\tbl_{ti})$ and $\mathcal{R}(\btbl) = \sum_{t=2}^{N-1} \sum_{i \in [R]} g(\tbl_{ti})$, and let $D$ be the feasible region of (\ref{problem: MAP on path}). 
\begin{proposition} \label{prop: upper bound}
Let $\bar{g}_{ti}^{(s)}(z) \defeq - \log  (\tbl_{ti}^{(s)}! ) + \alpha_{ti}^{(s)} \cdot (z - \tbl_{ti}^{(s)})$, where $\alpha_{ti}^{(s)}$ is a real number which satisfies $-\log(\tbl_{ti}^{(s)}+1)\leq \alpha_{ti}^{(s)} \leq -\log \tbl_{ti}^{(s)}$.  
Then, the function $\bar{\mathcal{R}}^{(s)}(\btbl) \defeq \sum_{t=2}^{N-1} \sum_{i \in [R]} \bar{g}_{ti}^{(s)}(\tbl_{ti})$ satisfies $\bar{\mathcal{R}}^{(s)}(\btbl^{(s)}) = \mathcal{R}(\btbl^{(s)})$ and $\bar{\mathcal{R}}^{(s)}(\btbl) \geq \mathcal{R}(\btbl)$. 
\end{proposition}
Please see the Appendix for the proof. Proposition \ref{prop: upper bound} says that the function $\bar{\mathcal{R}}^{(s)}(\btbl)$ satisfies conditions (i) and (ii). 
Intuitively, $\bar{g}_{ti}^{(s)}(\btbl)$ is a tangent of $g$ at $n_{ti}$. 

We check condition (iii). 
We consider the MCFP instance constructed by Algorithm \ref{alg: graph construction}. Then, we slightly modify this instance by changing the cost function of edge $(u_{t, i}, w_{t, i})$ from $g(z) + h_{ti}(z)$ to $\bar{g}_{ti}^{(s)}(z) + h_{ti}(z)$ for $t=2, \ldots, N-1$, $i \in [R]$. It can be easily verified that the minimization problem of $\bar{\mathcal{P}}^{(s)}(\btbl) $ in $D$ is equivalent to solving this new MFCP instance. 
Because all the cost functions on edges satisfy discrete convexity in this new MCFP instance, we can solve it efficiently by applying existing algorithms for C-MCFP\@.
(The details are discussed in Section \ref{subsec: complexity}.)

From above arguments, we can construct an efficient optimization algorithm for the MAP inference problem (\ref{problem: MAP on path}) by applying the DCA framework to our problem. 
An overall view of the derived algorithm is given in Algorithm \ref{alg: MAP_DC}. 
The algorithm is guaranteed to terminate after a finite number of iterations because $\mathcal{P}(\btbl^{(s)})$ monotonically decreases and $D$ is a finite set.


\begin{algorithm}[t]
  \caption{DCA for solving problem (\ref{problem: MAP on path})} \label{alg: MAP_DC}
  \begin{algorithmic}[1]
    \STATE{$\btbl^{(1)} \leftarrow \bm{0}$}
    \FOR{$s = 1,2,\dots$}
      \FOR{$t=2, \ldots, N-1$ and $i \in [R]$}
        \STATE{replace the cost function of edge $(u_{t, i}, w_{t, i})$ in $\flowG$ by $\bar{g}_{ti}^{(s)}(z) + h_{ti}(z)$}
      \ENDFOR
      \STATE{$\btbl^{(s+1)}\leftarrow$ solve C-MCFP in $\flowG$}
      \IF{$\mathcal{P}(\btbl^{(s)}) = \mathcal{P}{(\btbl^{(s+1)})}$}
        \STATE{\textbf{return} $\btbl^{(s)}$}
      \ENDIF
    \ENDFOR
  \end{algorithmic}
\end{algorithm}

\begin{table*}[t] 
  \centering
  \caption{Attained objective functions in synthetic instances. For each setting, we generated 10 instances and average values are shown. The smallest value is highlighted for each setting. U and D mean the ``uniform'' and ``distance'' potential setting, respectively. } \label{table: synthetic result}
  \vspace{0.1in}
  \scriptsize
  \begin{tabular}{cl|ccc|ccc|ccc} \hline
       & $M$ & \multicolumn{3}{c|}{$10^1$} & \multicolumn{3}{c|}{$10^2$} & \multicolumn{3}{c}{$10^3$}   \\ \hline
       & $R$ & 10 & 20 & 30 & 10 & 20 & 30 & 10 & 20 & 30 \\ \hline
       & Proposed (L) & \bf{-9.97e+01} & \bf{-8.90e+01} & \bf{-8.74e+01} & \bf{-1.11e+03} & \bf{-1.19e+03} & \bf{-1.22e+03} & \bf{-1.07e+04} & \bf{-1.31e+04} & \bf{-1.40e+04} \\
       U & Proposed (M) & -9.81e+01 & -8.90e+01 & \bf{-8.74e+01} & -1.11e+03 & -1.19e+03 & -1.22e+03 & -1.07e+04 & -1.31e+04 & -1.40e+04 \\
        & Proposed (R) & -9.64e+01 & -8.76e+01 & \bf{-8.74e+01} & -1.11e+03 & -1.18e+03 & -1.21e+03 & -1.07e+04 & -1.31e+04 & -1.40e+04 \\
       & NLBP & -7.19e+01 & -7.01e+01 & -7.01e+01 & -1.08e+03 & -9.87e+02 & -9.02e+02 & -1.07e+04 & -1.30e+04 & -1.37e+04 \\ \hline
       & Proposed (L) & \bf{3.35e-01} & \bf{5.00e-01} & \bf{5.00e-01} & \bf{-5.48e+01} & \bf{-3.03e+01} & \bf{-1.18e+01} & \bf{-5.83e+00} & \bf{-9.06e+02} & \bf{-1.01e+03} \\
       D & Proposed (M) & \bf{3.35e-01} & \bf{5.00e-01} & \bf{5.00e-01} & -5.43e+01 & -2.91e+01 & -1.14e+01 & -5.82e+00 & -9.06e+02 & -1.01e+03 \\
        & Proposed (R) & \bf{3.35e-01} & \bf{5.00e-01} & \bf{5.00e-01} & -5.39e+01 & -2.89e+01 & -1.06e+01 & -5.80e+00 & -9.06e+02 & -1.01e+03 \\
       & NLBP & 3.20e+01 & 4.56e+01 & 5.28e+01 & -1.38e+01 & 1.77e+02 & 3.25e+02 & 1.20e+00 & -8.02e+02 & -5.31e+02 \\ \hline
\end{tabular}
\scriptsize
\end{table*}

\subsection{Time complexity of one iteration} \label{subsec: complexity}
We analyze the time complexity of one iteration of the proposed method (Lines 3--7 in Algorithm \ref{alg: MAP_DC}). 
The computation bottleneck is solving C-MCFP in $\flowG$ (Line 5). 
There are several algorithms to solve C-MCFP and time complexity varies depending on which one is adopted.
In this paper, we consider two typical methods, Successive Shortest Path (SSP) and Capacity Scaling (CS) \cite{Ahuja1993}.  

For C-MCFP with Graph $\flowG = (\flowV, \flowE)$ and total flow $M$, the time complexity of SSP is $O(M |\flowE| \log |\flowV|)$ and that of CS is $O(|\flowE|^2 \log |\flowV| \log M)$. 
Because $|\flowV| = O(NR)$ and $|\flowE| = O(NR^2)$ in our problem, time complexity in one iteration is $O(MNR^2 \log (NR))$ when SSP is applied and $O(N^2R^4 \log (NR) \log M)$ when CS is applied.  
This result implies that each method has its own advantages and disadvantages: SSP has small computation complexity for $N$ and $R$, while CS has small computation complexity for $M$. This difference is confirmed empirically in Section \ref{subsec: synthetic}.

\section{Experiments} \label{sec: experiments}
We perform experiments to evaluate the effectiveness of the proposed method using synthetic and real-world instances. 
All experiments are conducted on a 64-bit macOS machine with Intel Core i7 CPUs and 16 GB of RAM. 
All algorithms are implemented in C++ (gcc 9.1.0 with -O3 option). 

\subsection{Synthetic Instances} \label{subsec: synthetic}
\paragraph{Settings.}
We solve randomly generated synthetic instances of the MAP inference problem (\ref{problem: MAP on path}).
We fix $T$ to 5 and vary the values of $R$ and $M$. 
The input observation $y_{ti}$ is independently drawn from uniform distribution on the set of integers $\{1, 2, \ldots, 2 \cdot \lfloor \frac{M}{R} \rfloor \}$. 
As the noise distribution, we use Gaussian distribution 
$p_{ij}(y_{ij}|t_{ij}) \propto \exp \big( \frac{-(y_{ij} - t_{ij})^2}{100}\big)$. 
We use two types of potential functions as follows. 
(1) \textbf{uniform}. $\phi_{tij}$ is independently drawn from uniform distribution on the set of integers $\{1, 2, \ldots, 10 \}$. 
(2) \textbf{distance}. We set $\phi_{tij} = \frac{1}{|i-j+1|}$. This potential models the movement of individuals in one-dimensional space: the state indices $i$ and $j$ represent coordinates in the space, and the closer the two points are, the more likely are movements between them to occur.

\paragraph{Proposed Method.}
To construct surrogate functions in the proposed method, we can choose arbitrary $\alpha_{ti}^{(s)}$ which satisfies the condition $-\log(\tbl_{ti}^{(s)}+1)\leq \alpha_{ti}^{(s)} \leq -\log \tbl_{ti}^{(s)}$ (see Proposition \ref{prop: upper bound}). 
To investigate the influence of the choice of $\alpha_{ti}^{(s)}$, we try three strategies to decide $\alpha_{ti}^{(s)}$:
(1) $\alpha_{ti}^{(s)} = -\log(\tbl_{ti}^{(s)})$,
(2) $\alpha_{ti}^{(s)} = -\frac{1}{2} (\log(\tbl_{ti}^{(s)}) + \log(\tbl_{ti}^{(s)}+1) )$,
(3) $\alpha_{ti}^{(s)} = -\log(\tbl_{ti}^{(s)}+1)$.
We call them Proposed (L), Proposed (M), Proposed (R), respectively. 

\paragraph{Compared Method.}
As the compared method, we use Non-Linear Belief Propagation (NLBP) \cite{Sun2015}, which is a message-passing style algorithm to the solve approximate MAP inference problem derived by applying Stirling's approximation and continuous relaxation. 
Because the output of NLBP is not integer-valued and $\log (z!)$ is defined only if $z$ is an integer, we cannot calculate the objective function of (\ref{problem: MAP on path}) directly. 
To address this, we calculate it by replacing the term $\log (z!)$ by linear interpolation of $\log (\lfloor z \rfloor!)$ and $\log (\lceil z \rceil!)$, which is given by $(\lceil z \rceil - z) \cdot \log (\lfloor z \rfloor!) + (z - \lfloor z \rfloor) \cdot \log (\lceil z \rceil!)$. 
Note that although there are various algorithms to solve the approximate MAP inference problem (see Section \ref{subsec: related MAP}), the objective function values attained by these algorithms are the same.  
This is because the approximate problem is a convex optimization problem \cite{Sheldon2013a}.

\paragraph{Comparison of attained objective values.} \label{subsec: synthetic objective}
The results are shown in Table \ref{table: synthetic result}. 
We generated 10 instances for each parameter setting and determined the average of attained objective function values. 
Because the objective function $\mathcal{P}(\btbl)$ is equal to $-\log \Pr(\btbl|\by) + \mathrm{const. }$, $\mathcal{P}(\btbl)$ takes both positive and negative values, and the difference of the objective function values is essential; when $\mathcal{P}(\btbl_1) - \mathcal{P}(\btbl_2) = \delta$, $\Pr(\btbl_1|\by) = \exp (-\delta) \cdot \Pr(\btbl_2|\by)$ holds. 

All the proposed methods consistently have smaller objective function values than the compared method.
The difference tends to be large when $R$ is large and $M$ is small. This would be because small values appear in the contingency table more frequently when $R$ is large and $M$ is small, and the effect of the inaccuracy of Stirling's approximation becomes larger.
Among three proposed methods, there was not much difference in obtained objective function values, but Proposed (L) was found to consistently achieve slightly smaller objective function values than others.

\begin{figure}
  \centering
  \fbox{\includegraphics[width=0.75\columnwidth]{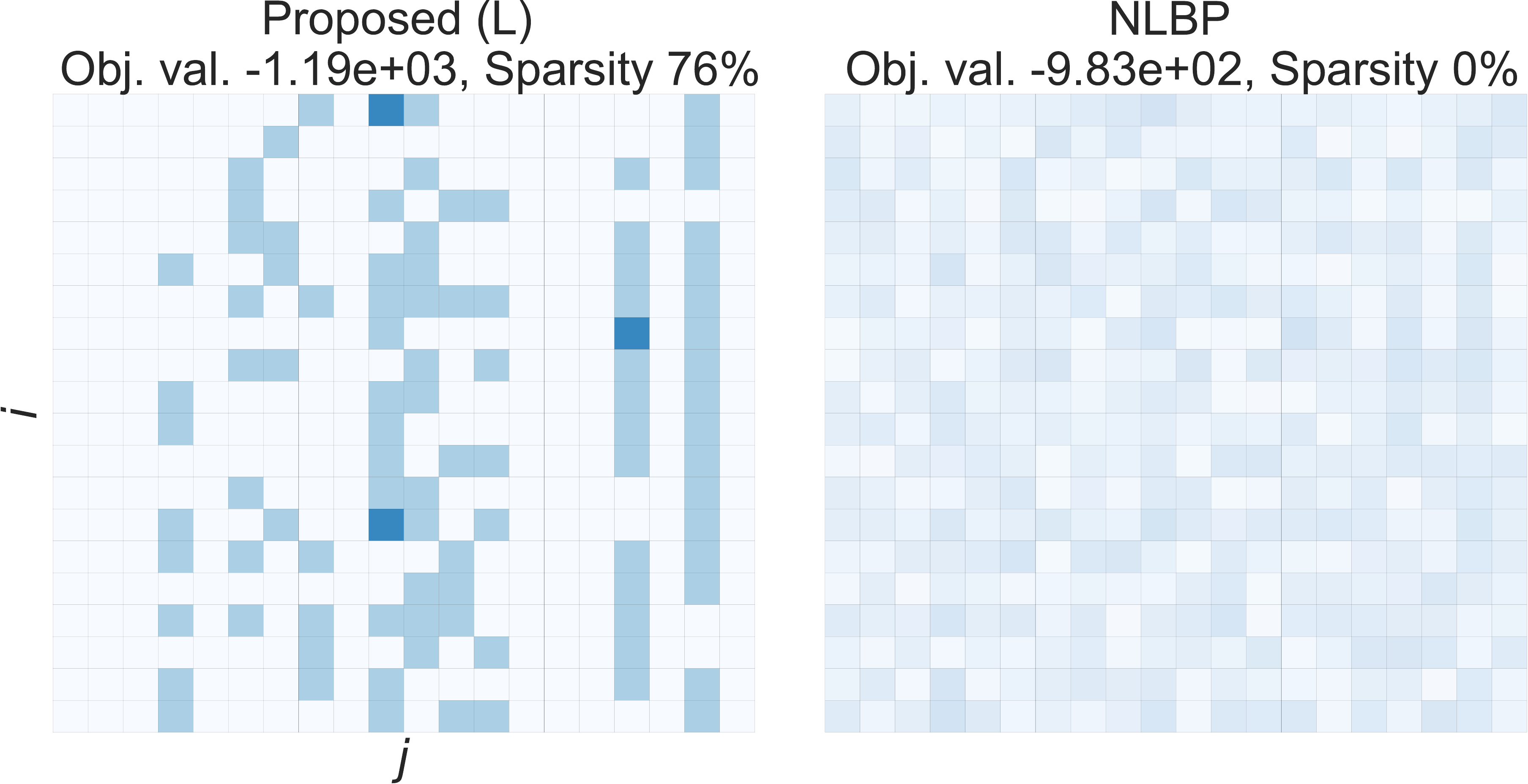}}\par
  \fbox{\includegraphics[width=0.75\columnwidth]{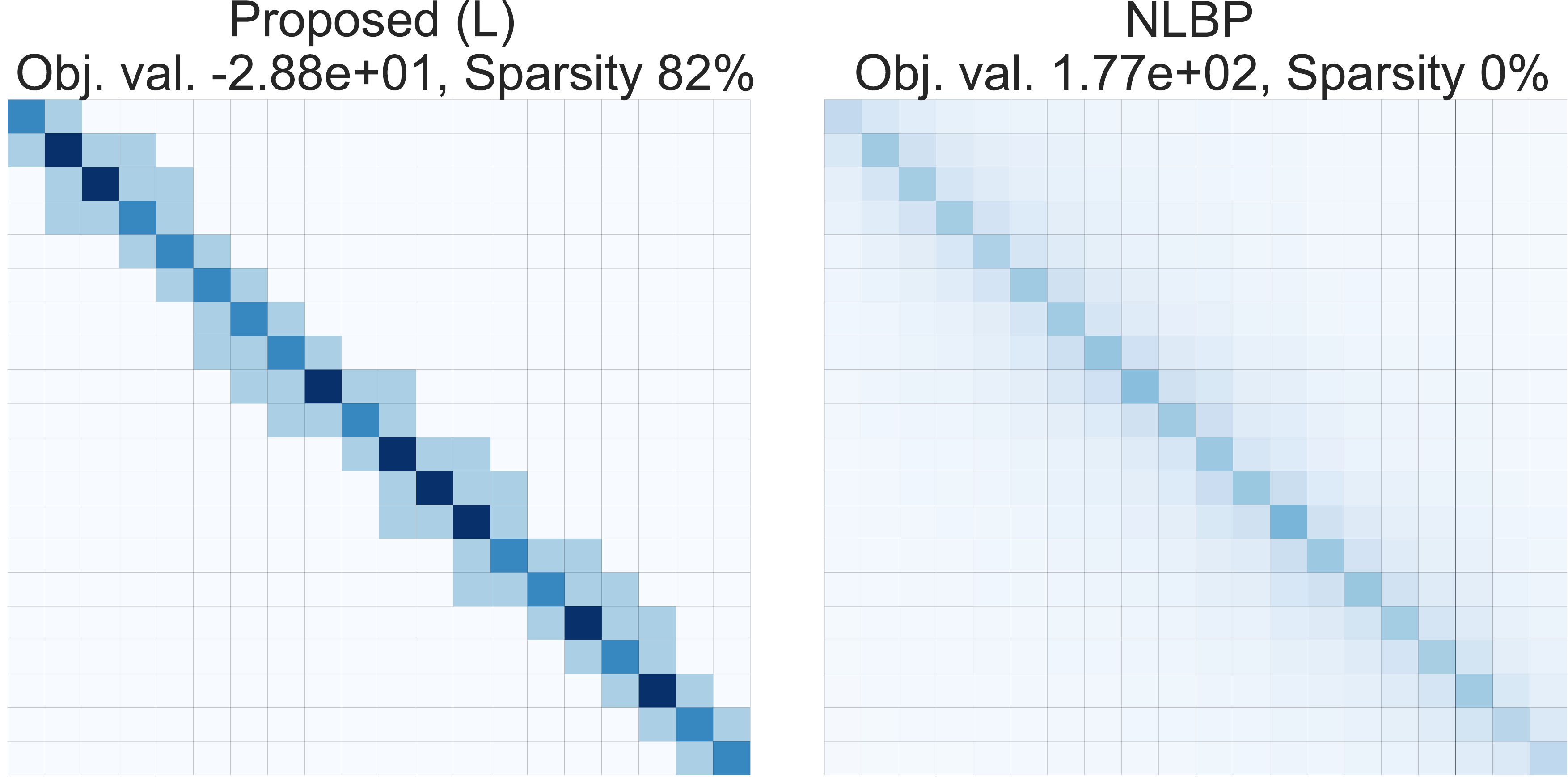}}%
  \caption{Comparison of solutions yielded by proposed method (L) and NLBP. 
  We solve two instances with $R=20$, $M=10^2$ and uniform potential (top) and $R=20$, $M=10^2$ and distance potential (bottom).
  The obtained edge contingency table $\tbl_{1ij}$ is presented as a matrix heatmap with maximum value of color map 3.  
  For both potential settings, the solution by the proposed method is quite sparse while that by NLBP contains a lot of non-zero elements. } \label{fig: heatmap} 
  \label{fig: sparsity}
\end{figure}

\paragraph{Characteristics of the output solution.} 
To compare the characteristics of solutions obtained by proposed (L) and NLBP, we solve two instances (1) $R=20$, $M=10^2$, uniform potential (2) $R=20$, $M=10^2$, distance potential by each method. 
Obtained edge contingency tables $\tbl_{1ij}$ are shown in Figure \ref{fig: heatmap} as heat maps. 
For both potential settings, proposed method outputs sparse solutions while the solutions by NLBP are blurred and contain a lot of non-zero elements. 
This difference is quantified by ``sparsity'', which is calculated by $1.0-$ (\# of non-zero ($>10^{-2}$) elements)/(\# of elements): sparsity of the output of proposed (L) is 76\% (uniform potential), 83\% (distance potential), while the sparsity of the output of NLBP is 0\% in both settings. 
This is caused by its application of continuous relaxation and the inaccuracy of Stirling's approximation around 0. 
In additional experiments, we observed that the output of the two methods become closer as $M$ increases. For more details, please see the Appendix.

\begin{table*}
\centering
  \caption{Attained objective functions for real-world instances. For each setting, we generated 10 instances and average values are shown. The smallest value is highlighted for each setting. } \label{table: real-world result}
  \vspace{0.1in}
\scriptsize
  \begin{tabular}{l|cc|cc|cc} \hline
       $M$ & \multicolumn{2}{c|}{$10^1$} & \multicolumn{2}{c|}{$10^2$} & \multicolumn{2}{c}{$10^3$}   \\ \hline
       $R$ & 56 & 208 & 56 & 208 & 56 & 208  \\ \hline
       Proposed (L) & \bf{2.30e+00} & \bf{2.30e+00} & \bf{-1.26e+03} & \bf{2.31e+01} & \bf{-2.22e+04} & \bf{-2.22e+04} \\
       Proposed (M) & \bf{2.30e+00} & \bf{2.30e+00} & -1.25e+03 & \bf{2.31e+01} & -2.22e+04 & -2.22e+04 \\
       Proposed (R) & \bf{2.30e+00} & \bf{2.30e+00} & -1.20e+03 & \bf{2.31e+01} & -2.20e+04 & -2.21e+04 \\
       NLBP & 2.76e+02 & 3.83e+02 & 1.91e+03 & 3.85e+03 & -1.47e+04 & 1.73e+04\\ \hline
\end{tabular}
\end{table*}

\paragraph{Comparison of computation time.}
We compare the computation time of each algorithm. 
As explained in Section \ref{subsec: complexity}, we can choose an arbitrary C-MCFP algorithm as the subroutine in the proposed method and the time complexity varies depending on the choice.
We compare proposed (L) with SSP, proposed (L) with CS, and NLBP\@. 

\begin{figure}
  \includegraphics[width=0.95\columnwidth]{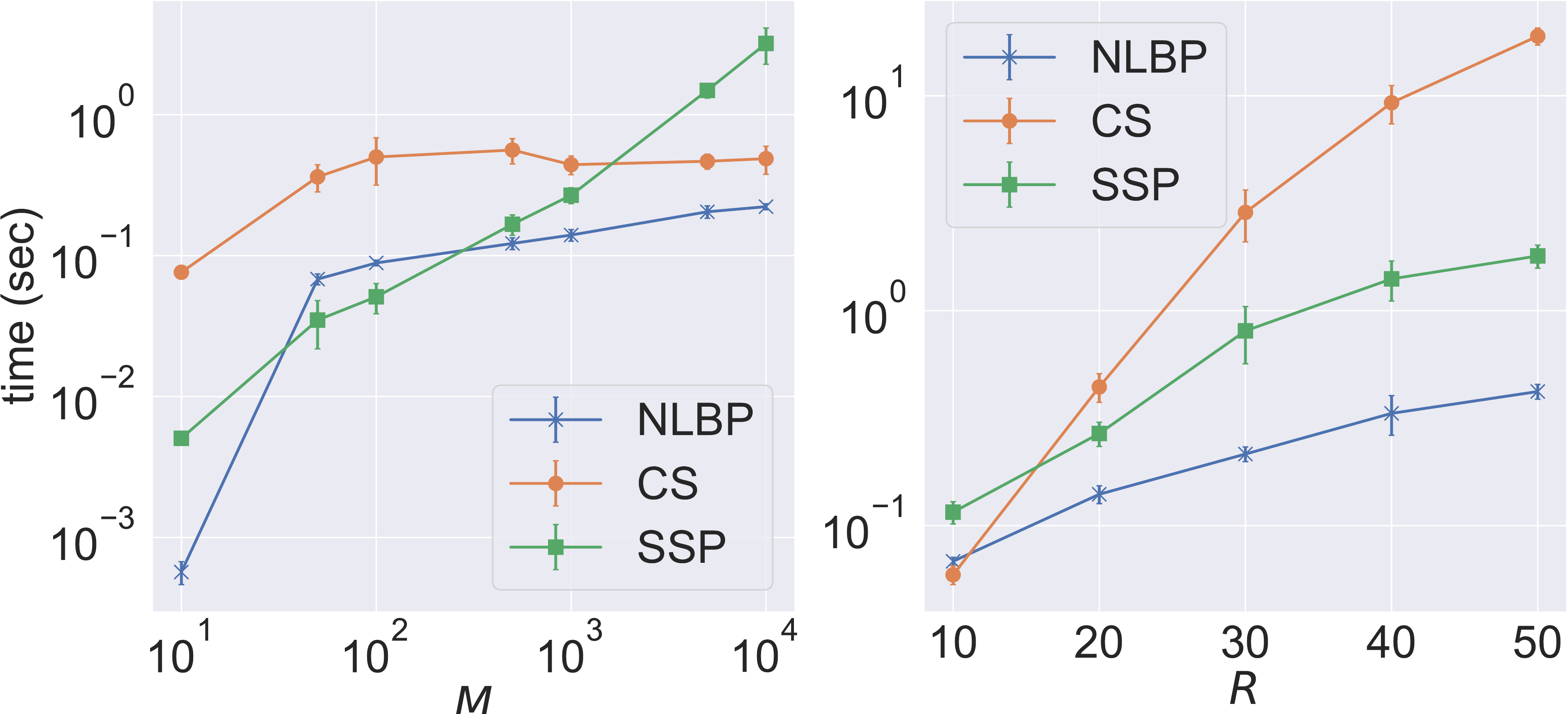}
  \vspace{-0.05in}
  \caption{The average computation time of each algorithm. The values are averages of 10 synthetic instances when $R$ is fixed to 20 (left) an $M$ is fixed to $10^3$ (right). $N$ is set 5 and uniform potential is used. The error bars represent standard deviations. } \label{fig: speed} 

  \par\vspace{0.2in}
  \includegraphics[width=0.95\columnwidth]{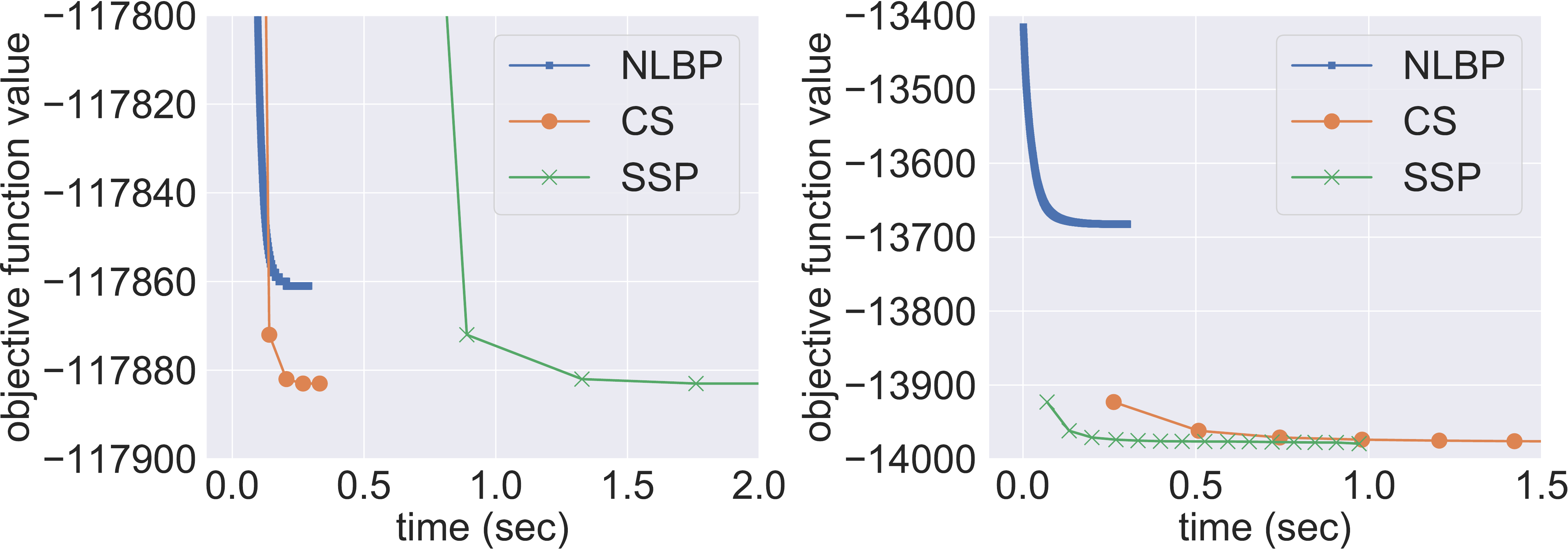}
  \vspace{-0.05in}
  \caption{The relationship between running time and objective function value. The left figure shows the result of an instance of $R = 20, M=10^4$ and uniform potential, and right figure shows that of $R = 30, M=10^3$ and uniform potential.} \label{fig: time vs obj} 
\end{figure}

Figure \ref{fig: speed} shows the relationship between input size and computation time, and
Figure \ref{fig: time vs obj} shows the relationship between running time and objective function value.
These results are consistent with the complexity analysis results in Section \ref{subsec: complexity};
SSP is efficient when $R$ is large but becomes inefficient when $M$ is large, and the converse is true for CS.
The results also suggest that it is important to choose the algorithm depending on the size parameter of the input.
The proposed method is not much worse than the existing method in terms of computation time by choosing an appropriate C-MCFP algorithm according to the size of the input. 
Appropriately chosen proposed methods attain the minimum of the existing method more quickly; proposed methods take a lot of time to achieve a smaller objective function value than the minimum of the existing method. 

\subsection{Real-world Instances}
We conduct experiments using real-world population datasets. 
The datasets are generated from 8694 car trajectories collected by a car navigation application in the Greater Tokyo area, Japan \footnote{We use the car data collected by the smartphone car navigation application of NAVITIME JAPAN Co., Ltd. (http://corporate.navitime.co.jp/en/)}. 
We randomly sample $M\ (M=10^1, 10^2, 10^3)$ trajectories from this data and create aggregated population data of each area at fixed time intervals.
The areas are decided by dividing the targeted geospatial space into fixed size grid cells. 
The grid size is set to 10km $\times$ 10km ($R = 8 \times 7 = 56$) and 5km $\times$ 5km ($R = 16 \times 13 = 208$), and time interval is 60 minutes ($N=24$). 
We construct the potential $\phi_{tij} = \frac{1}{1+\dist(i, j)}$, where $\dist(i, j)$ is the Euclidean distance between the centers of cell $i$ and cell $j$ in the grid space. 
We created 10 instances by random sampling and averaged the attained objective function values for each setting. 

Table \ref{table: real-world result} shows the results. 
The trends are similar to those found in experiments on synthetic instances: 
(i) Proposed methods consistently attain smaller objective values than the existing method. 
(ii) Superiority of the proposed method increase when $R$ is large and $M$ is small. 
(iii) Proposed (L) attains the best score among the proposed methods, but the difference is relatively small.

\section{Related Work} \label{sec: related work}
\subsection{MAP inference for CGMs} \label{subsec: related MAP}
Several methods have been proposed for the MAP inference of CGMs, but most of them take the approach of solving the approximate problem \cite{Sheldon2013a}, which is derived by applying Stirling's approximation and continuous relaxation.
For example, the interior point method \cite{Sheldon2013a}, projected gradient descent \cite{Vilnis2015}, message passing \cite{Sun2015} and  Sinkhorn-Knopp algorithm \cite{Singh2020} have been used to solve the approximate problem. 
In particular, \cite{Nguyen2016} proposes a method to use DCA to solve this approximate problem. 
Although this approach is similar to our proposal in that it uses DCA, the purpose of applying DCA is totally different: our focus is to solve the MAP inference problem without using any approximation or continuous relaxation.


One of the few exceptions is the method proposed in \cite{Akagi2020}, which solves the original MAP inference problem directly without using approximation.
Our method follows this line of research, but there are two major differences. First, their method can only be applied to CGM on a graph with two vertices, and thus applicability is very limited. Since our method is consistent with this method when applied to CGM on a graph with two vertices, our method can be regarded as a generalization of their method.
Second, their work assumes accurate observations and so does not mention how to handle observation noise.

\cite{Sheldon2007} solves related collective MAP inference problems on path graphs. 
The problems addressed in this paper are different from ours; their purpose is finding the most likely assignments of the entire variables for each individual, while our purpose is finding the most likely node and edge contingency tables. 
In their settings, non-linear terms in the log posterior probability vanish, and the MAP inference problem can be solved easily by linear optimization approaches. 

\subsection{Difference of Convex Algorithm (DCA)}
DCA, which is sometimes called Convex Concave Procedure \cite{Yuille2001}, is a framework to minimize a function expressed as the sum of a convex function and a concave function \cite{Le2018}. 
DCA was originally proposed as a method for optimization in continuous domains. 
DCA has been used in various machine learning fields, such as feature selection \cite{Le2015}, reinforcement learning \cite{Piot2014}, support vector machines \cite{Xu2017} and Boltzmann machines \cite{Nitanda2017}. 

Several studies have applied DCA to discrete optimization problems. 
This line of research is sometimes called discrete DCA \cite{Maehara2015a}. 
\cite{Narasimhan2005, Iyer2012} propose algorithms to minimize the sum of a submodular function and a supermodular function. 
This algorithm is generalized to yield the minimization of the sum of an M/L-convex function and an M/L-concave function \cite{Maehara2015a}, where M-convex function and L-convex function are classes of discrete convex functions \cite{Murota1998}. 
Although our work is closely related to these studies, it is not part of them. 
This is because our problem can be regarded as the minimization of the sum of two M-convex functions and a separable concave function, and this is not included in the class of functions dealt with in \cite{Maehara2015a}
\footnote{A separable convex function is both L-convex and M-convex, but a sum of two M-convex functions is neither M-convex nor L-convex.}. 

\section{Conclusion}
In this paper, we propose a non-approximate method to solve the MAP inference problem for CGMs on path graphs. 
Our algorithm is based on an MCFP formulation and application of DCA. 
In the algorithm, surrogate functions can be constructed in closed-form and minimized efficiently by C-MCFP algorithms. 
Experimental results show that our algorithm outperforms approximation-based methods in terms of quality of solutions.  

\bibliography{main.bib}
\bibliographystyle{for_arxiv}

\clearpage

\appendix

\twocolumn[
\centering
\LARGE
\vspace{1.2\baselineskip}
\textbf{Appendices}
\par
\vspace{1.5\baselineskip}
]

\section{Derivation of (\ref{problem: MAP on path})} \label{appendix: MAP derivation}
Because 
\begin{align}
\nu_t = 
\begin{dcases*}
1 & if $t=1, N$, \\
2 & otherwise 
\end{dcases*}
\end{align}
holds on path graphs, we have 
\begin{align}
  F(\btbl) &= \frac{M!}{Z^M} \cdot \frac{\prod_{t=2}^{N-1} \prod_{i \in [R]} \tbl_{ti}! }{\prod_{t=1}^{N-1} \prod_{i, j \in [R]} \tbl_{tij}!} 
  \cdot \prod_{t=1}^{N-1} \prod_{i, j \in [R]} \phi_{tij}^{\tbl_{tij}}
\end{align}
from (\ref{formula: F}). This gives
\begin{align}
  &{-\log F(\btbl)} - \log\Pr(\by|\btbl) \\
  ={}& {- \log M!} + M \log Z \\
  &- \sum_{t=2}^{N-1} \sum_{i \in [R]} \log \tbl_{ti}! + \sum_{t=1}^{N-1} \sum_{i, j \in [R]} \log \tbl_{tij}! \\
  &- \sum_{t=1}^{N-1} \sum_{i, j \in [R]} \tbl_{tij} \log \phi_{tij} - \sum_{t=1}^{N} \sum_{i \in [R]} \log p_{ti}(y|\tbl) \\
  ={}& \sum_{t=1}^{N-1} \sum_{i, j \in [R]} f_{tij}(\tbl_{tij})
  +\sum_{t=2}^{N-1} \sum_{i \in [R]} g(\tbl_{ti}) \\
  &+ \sum_{t=1}^{N} \sum_{i \in [R]} h_{ti}(\tbl_{ti}) + C,  
\end{align}
where $C$ is a constant. 
We can verify easily that the feasible region of problem (\ref{problem: MAP on path}) is $\dL_M^\dZ$ defined in (\ref{formula: L}). 

\section{Proofs}
\subsection{Proof of Proposition \ref{prop: MCFP}} \label{appendix: proof of MCFP}
\begin{proof}
  There is a one-to-one correspondence between a feasible solution to problem (\ref{problem: MAP on path}), $\btbl$, and a feasible solution to the MCFP instance constructed by Algorithm \ref{alg: graph construction}, $\bz$, under the relationship $\tbl_{ti} = z_{u_{t, i} w_{t,i}}$ and $\tbl_{tij} = z_{w_{t, i} u_{t+1, j}}$;
  the constraint $\sum_{i \in [R]} \tbl_{ti} = M$ is equivalent to the supply constraints at node $o$ and $d$, the constraint $\sum_{j \in [R]} \tbl_{tij} = \tbl_{ti}$ corresponds to the flow conservation rule at node $w_{t, i}$ and the constraint $\sum_{i \in [R]} \tbl_{tij} = \tbl_{i+1, j}$ corresponds to the flow conservation rule at node $u_{t+1, j}$. 
  Moreover, corresponding $\btbl$ and $\bz$ and have the same objective function value in problem (\ref{problem: MAP on path}) and the MCFP instance, respectively.
  These facts yield the Proposition. 
\end{proof}

\subsection{Proof of Proposition \ref{prop: convexity}} \label{appendix: proof of convexity}
\begin{proof}
The function $\log z!$ is a discrete convex function, since
\begin{align}
  &\log (z+2)! + \log z! - 2 \log (z+1)! \\
  ={}& \log(z+2) - \log(z+1) \geq 0. 
\end{align}
This yields that $f_{tij}(z) = \log z! - z \cdot \log \phi_{tij}$ is a discrete convex function and $g(z) = - \log z!$ is a discrete concave function. 
Because a univariate continuous convex function is also a discrete convex function, $h_{ti}(z) = - \log \left[ p_{ti}(y_{ti}|z) \right]$ is a discrete convex function from Assumption \ref{assumption: log concave}. 
\end{proof}

\subsection{Proof of Proposition \ref{prop: upper bound}} \label{appendix: proof of upper bound}
\begin{proof}
First, we show that 
\begin{align}
- \log (w!) +  \alpha \cdot (z-w)  \geq - \log (z!),\quad  \forall z \in \dZp \label{formula: tangent}
\end{align}
holds for arbitrary $w \in \dZp$, when $-\log(w+1) \leq \alpha \leq -\log w$. 
When $z \geq w$, 
\begin{align}
  &-\log (w!) + \alpha \cdot (z-w) + \log (z!) \\
  ={}& \sum_{k=w+1}^z (\alpha + \log k) \geq 0
\end{align}
holds because $\alpha + \log(w+1) \geq 0$. 
When $z < w$, 
\begin{align}
  &-\log (w!) + \alpha \cdot (z-w) + \log (z!) \\
  ={}& \sum_{k=z+1}^w (-\alpha - \log k) \geq 0
\end{align}
holds because $- \alpha -\log w \geq 0$. Thus, inequality (\ref{formula: tangent}) holds. 

Substituting $w = \tbl_{ti}^{(s)}$ in (\ref{formula: tangent}), we get $\bar{g}_{ti}^{(s)}(z) \geq g (z)$ for all $z \in \dZp$. This yields 
\begin{align}
  \bar{\mathcal{R}}^{(s)}(\btbl) 
  = \sum_{t=2}^{N-1} \sum_{i=1}^{R} \bar{g}_{ti}^{(s)}(\tbl_{ti}) 
  \geq \sum_{t=2}^{N-1} \sum_{i=1}^{R} g (\tbl_{ti}) = \mathcal{R}(\btbl). 
\end{align}
Furthermore, since $\bar{g}_{ti}^{(s)}(\tbl_{ti}^{(s)}) = g (\tbl_{ti}^{(s)})$ from simple calculation, we get $\bar{\mathcal{R}}^{(s)}(\btbl^{(s)}) = \mathcal{R}(\btbl^{(s)})$. 
\end{proof}

\section{Additional experimental results} 
\subsection{The Characteristics of the solutions}\label{appendix: additional solution}

\begin{figure*}
  \centering
  \fbox{\includegraphics[width=0.90\columnwidth]{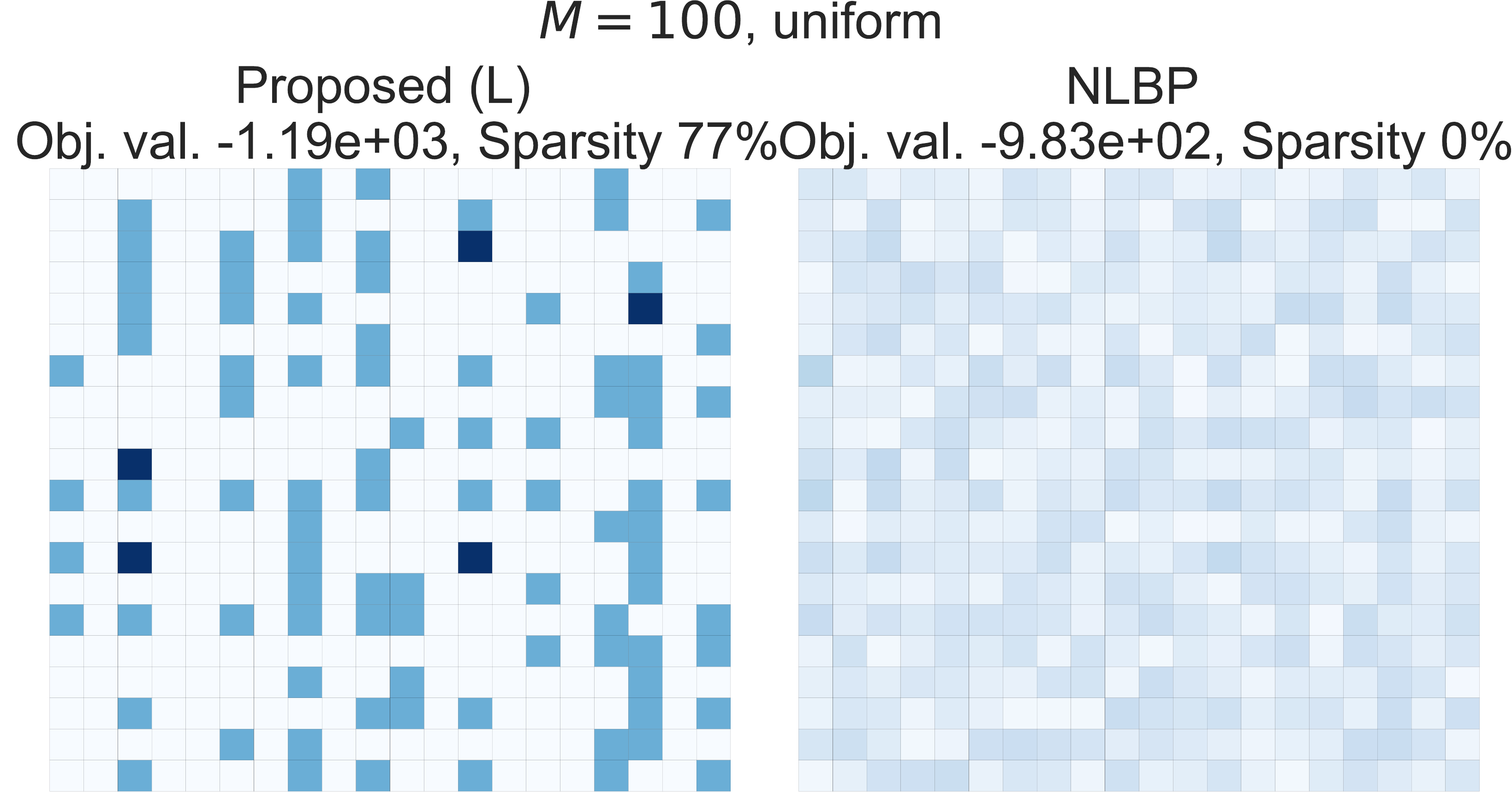}}
  \fbox{\includegraphics[width=0.90\columnwidth]{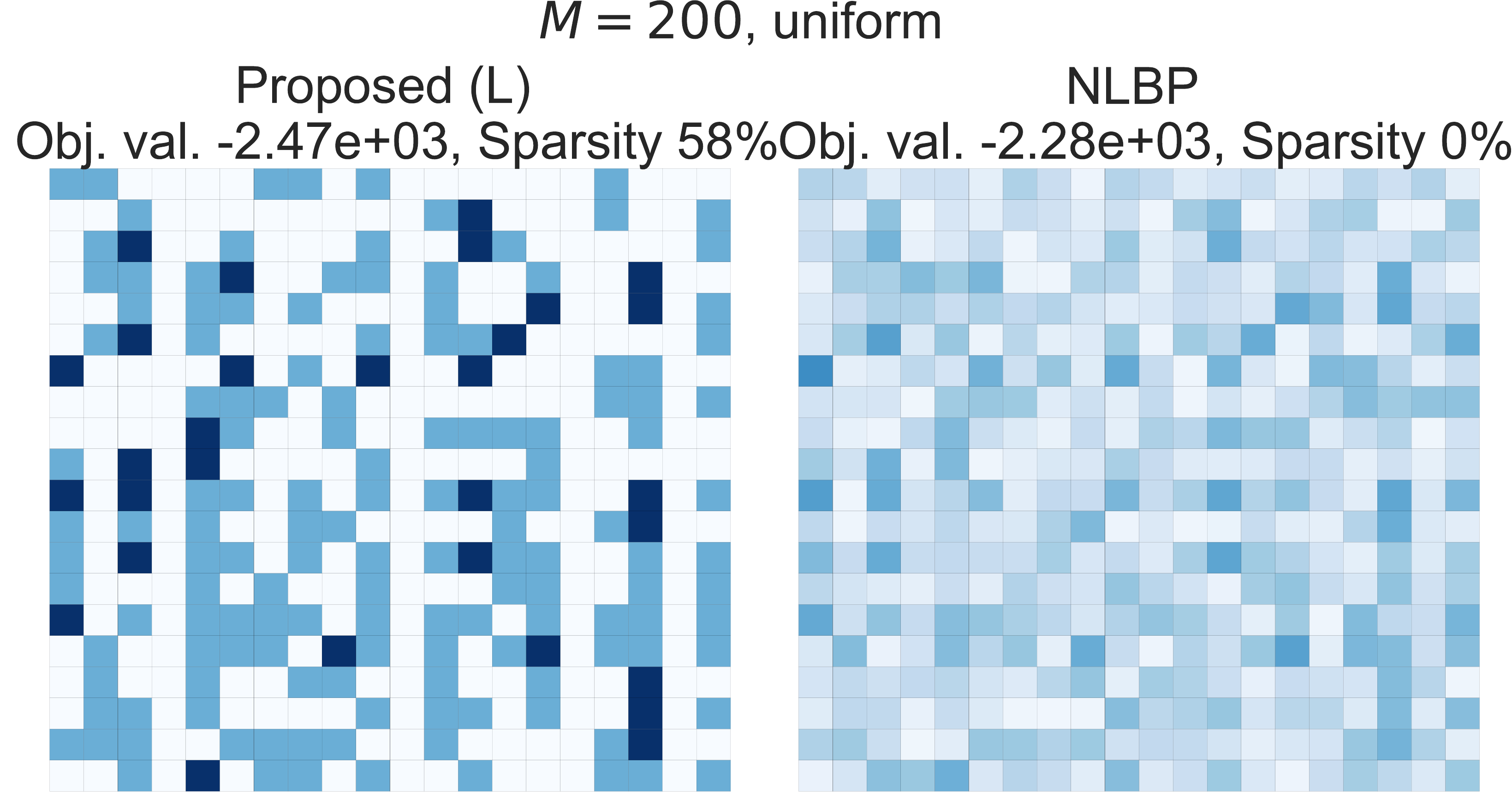}}\par
  \fbox{\includegraphics[width=0.90\columnwidth]{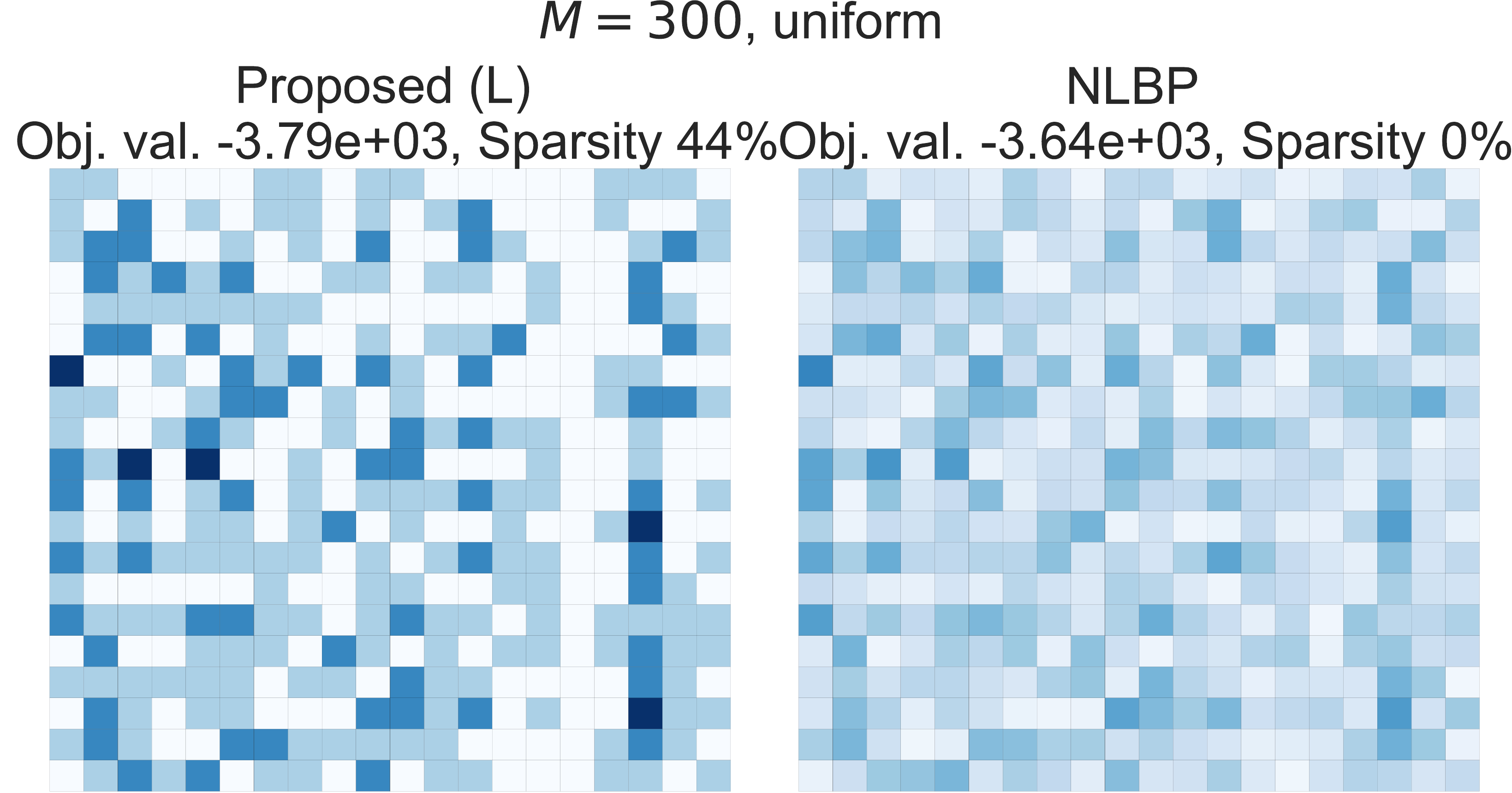}}
  \fbox{\includegraphics[width=0.90\columnwidth]{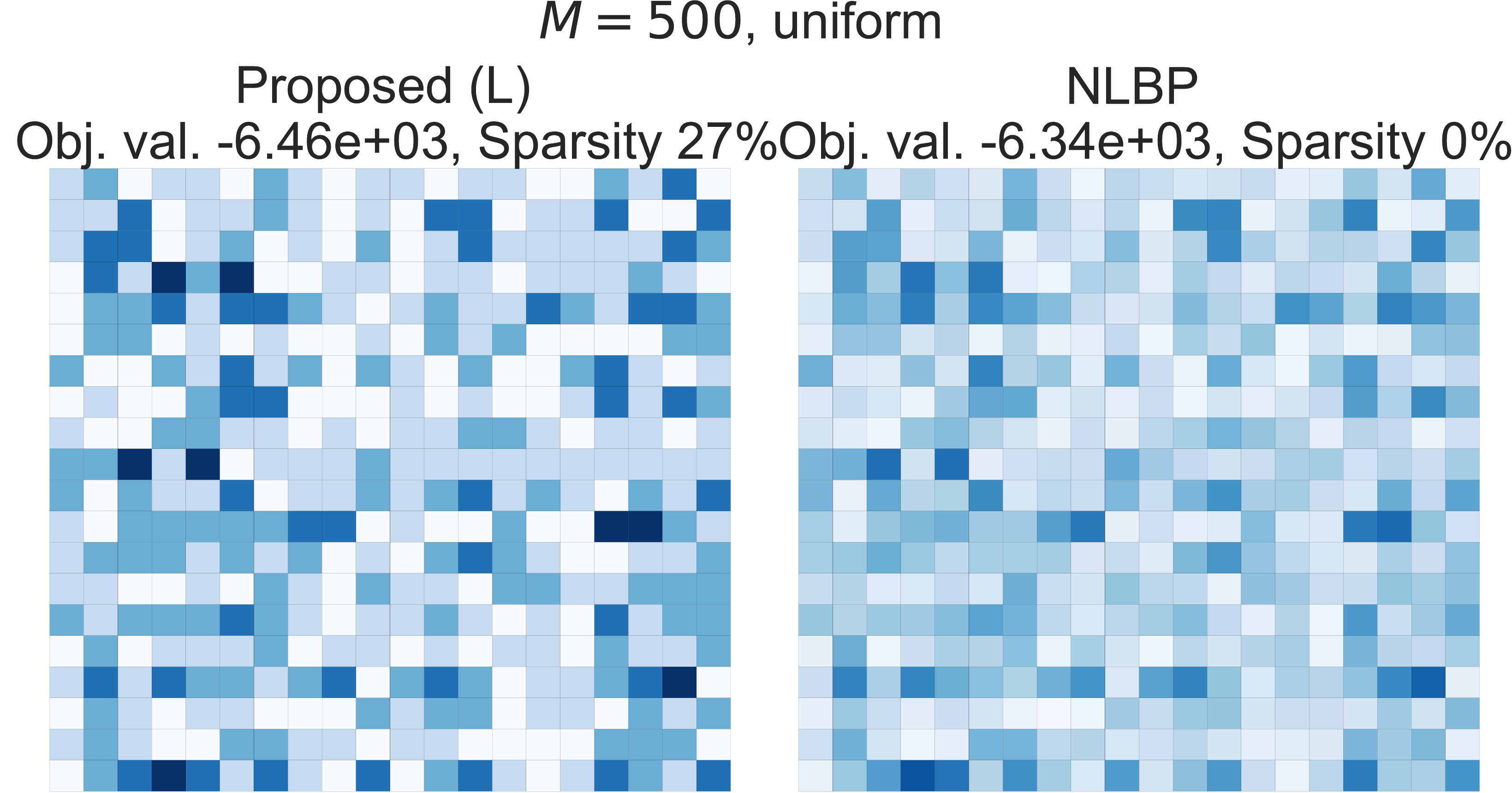}}\par
  \caption{Comparison of solutions yielded by the proposed method (L) and NLBP when $R = 20$ and the uniform potential is used. } \label{fig: heatmap uniform}
\end{figure*}

\begin{figure*}
  \centering
  \fbox{\includegraphics[width=0.90\columnwidth]{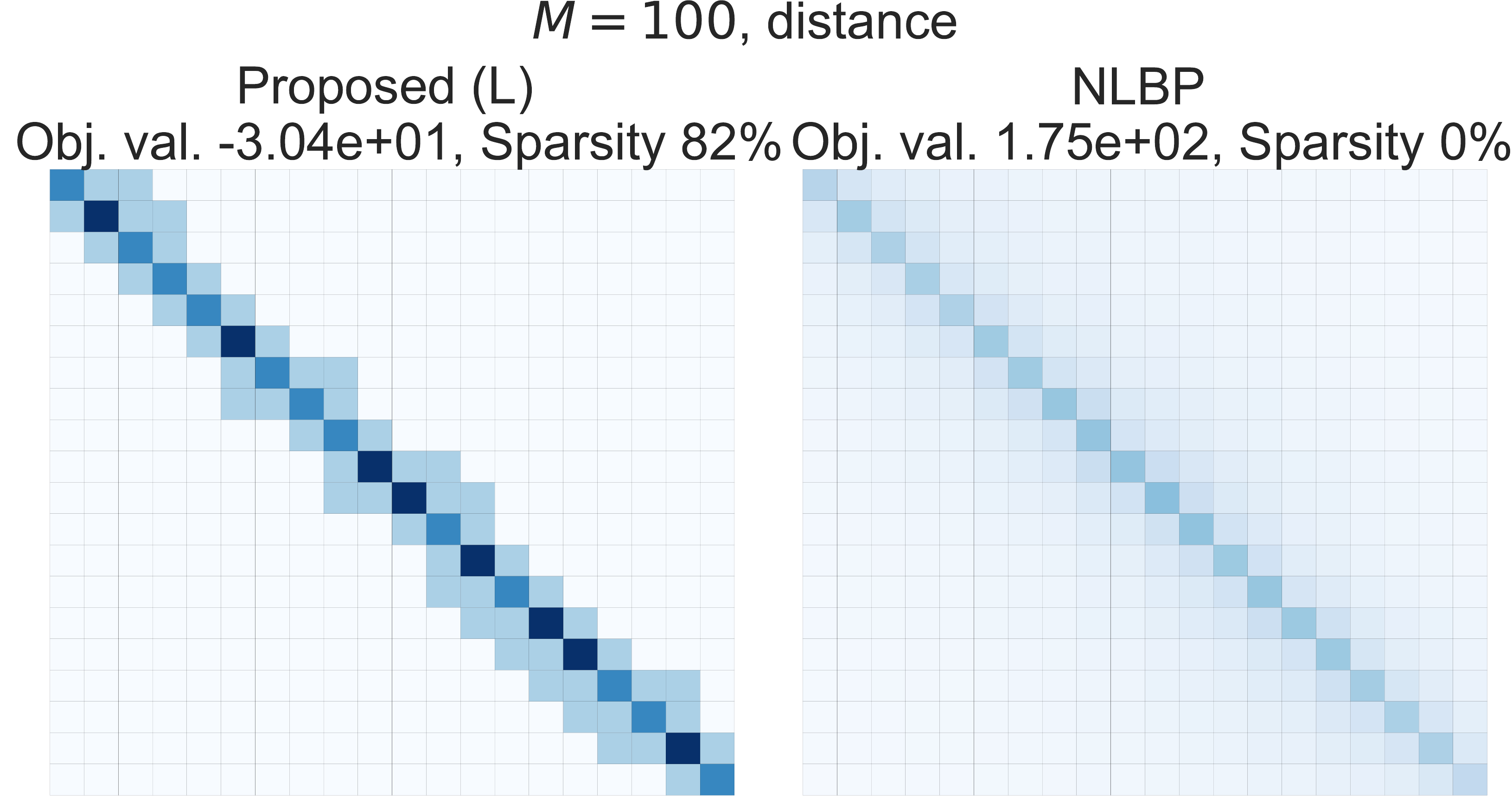}}
  \fbox{\includegraphics[width=0.90\columnwidth]{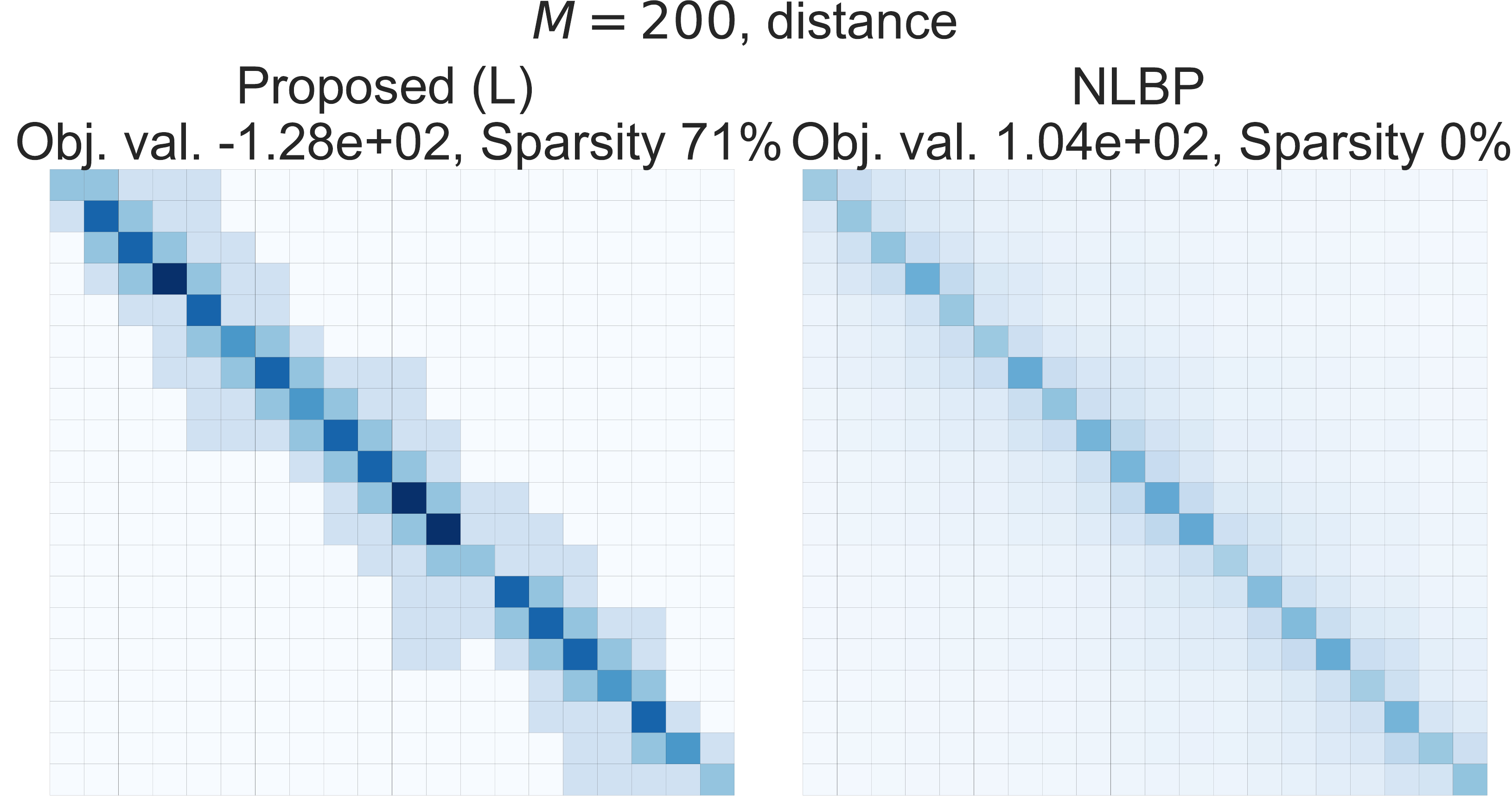}}\par
  \fbox{\includegraphics[width=0.90\columnwidth]{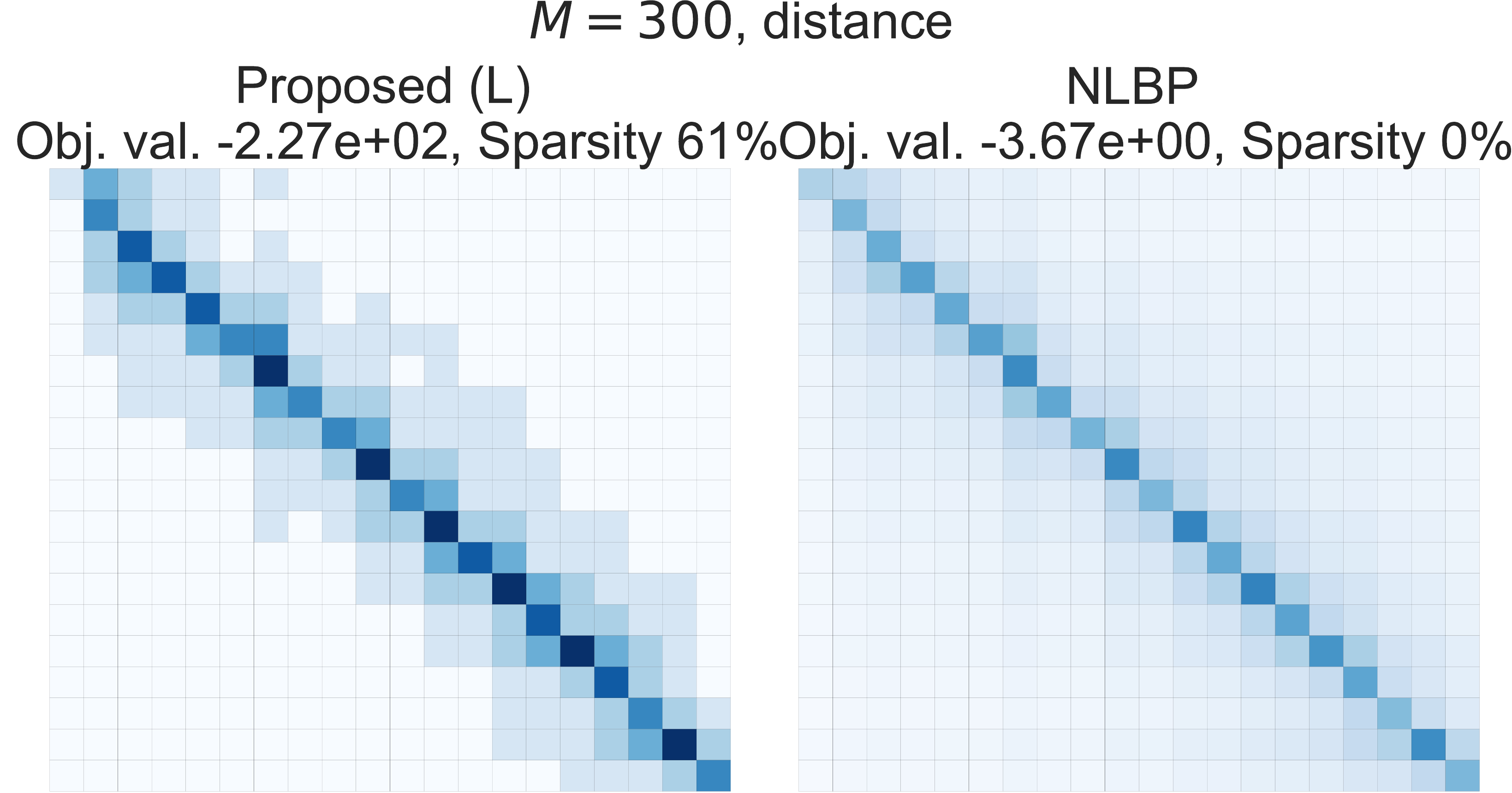}}
  \fbox{\includegraphics[width=0.90\columnwidth]{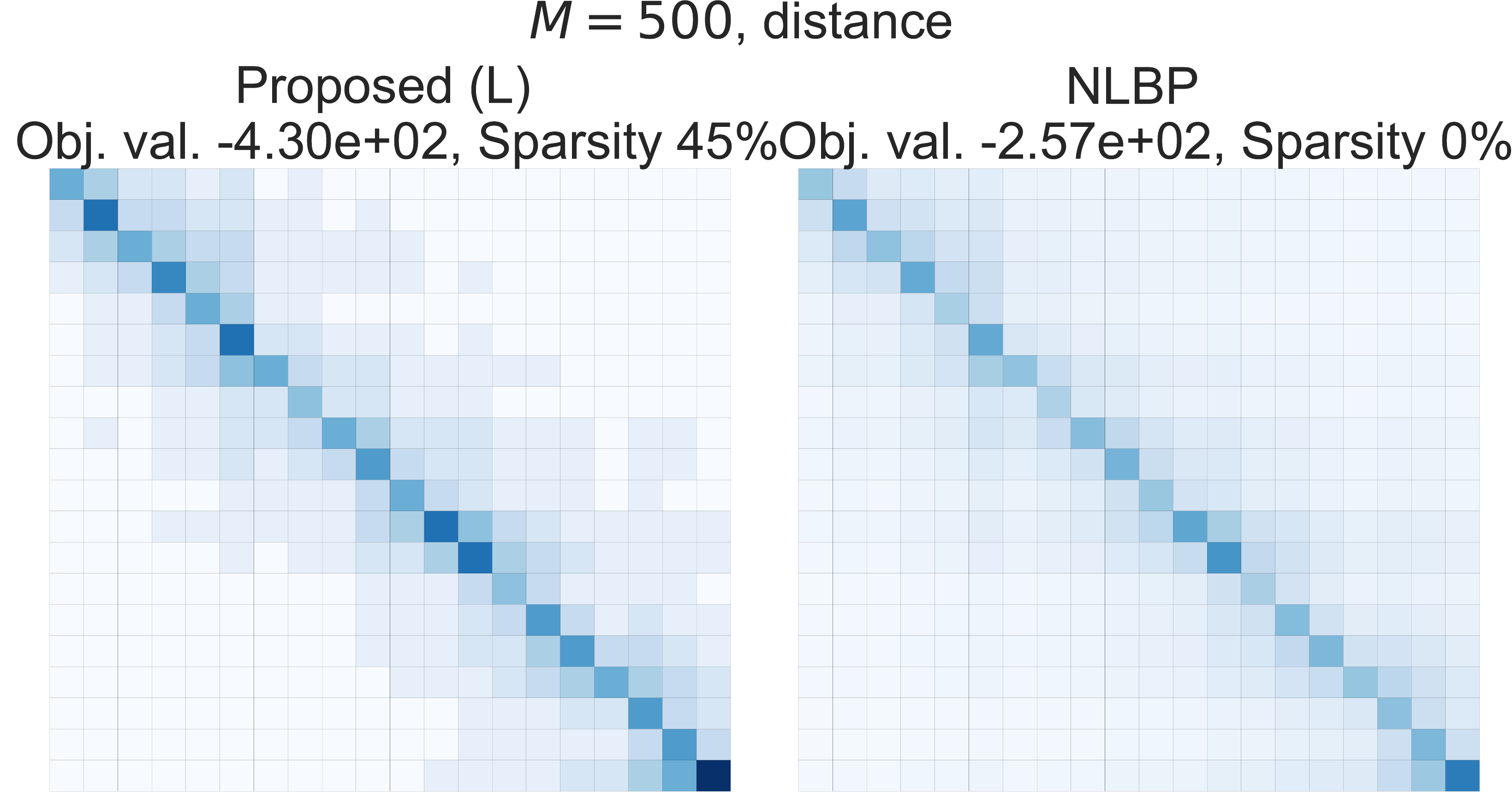}}\par
  \caption{Comparison of solutions yielded by the proposed method (L) and NLBP when $R = 20$ and the distance potential is used. } \label{fig: heatmap distance} 
\end{figure*}

We run the same experiments as \cref{fig: heatmap}
varying the value of $M$. The results are shown in \cref{fig: heatmap uniform,fig: heatmap distance}.
The outputs of the two methods are totally different when $M$ is small, and they get closer as $M$ increases. This is owing to the nature of Stirling's approximation $\log x! \approx x \log x - x $; it is inaccurate especially when $x$ is small.

\subsection{Histogram interpolation} \label{appendix: histogramu interpolation}
As an application of MAP inference of CGMs on path graphs, we can interpolate the time series of histograms between given two histograms. 
In this section, we show experimental results on this application and discuss the differences between the output of the proposed method and that of the existing method. 

\subsubsection{Settings}
First, we briefly explain how to realize interpolation between two histograms by MAP inference of CGMs on path graphs.
Suppose we are given histogram $\bheta_1 \defeq [\eta_{11}, \ldots, \eta_{1R}]$ at time $1$ and the histogram $\bheta_N \defeq [\eta_{N1}, \ldots, \eta_{NR}]$ at time $N$. The interpolated histogram $\bheta_t$ at time $t\ (= 2, \ldots, N-1)$ is calculated by the following procedure.

\begin{enumerate}
  \item Consider a CGM on a path graph with $N$ vertices.
  \item Let $\by_1 = \bheta_1$ and $\by_N = \bheta_N$.
  \item $\by_t \ (t= 2, \ldots, N-1)$ is treated as a \emph{missing value}. This can be achieved by setting $h_{ti}(z) = 0 \ (t=2, \ldots, N-1,\ i \in [R])$ in the objective function of the problem (\ref{problem: MAP on path}).
  \item Find a solution $\bn^*$ to the MAP inference problem under an appropriate potential $\bphi$.
  \item Obtain an interpolation result by $\eta_{ti} = \tbl^*_{ti} \ (t= 2, \ldots, N-1,\ i \in [R])$.
\end{enumerate}

In our experiment, we consider a grid space of size $5 \times 5 = 25 \ (= R)$ and a histogram $\bheta \defeq [\eta_1, \ldots, \eta_R]$ with a value $\eta_i$ for each cell $i \ (=1, \ldots, R)$.
To get interpolation results which consider the geometric structure defined by Euclidean distance in grid space, we set the potential $\phi_{tij} = \exp ( -( (x_i - x_j)^2 + (y_i - y_j)^2 ) )$, where $(x_i, y_i)$ is the two-dimensional coordinate of the center of cell $i$ in the grid space.
We set $N = 6$ and use Gaussian distribution $p_{ij}(y_{ij}|\tbl_{ij}) \propto \exp ( -5 (y_{ij} - \tbl_{ij})^2 )$ for the noise distributions at $t=1, N$.

\subsubsection{Results}
\begin{figure*}
  \centering
    \fbox{
      \includegraphics[width=1.5\columnwidth]{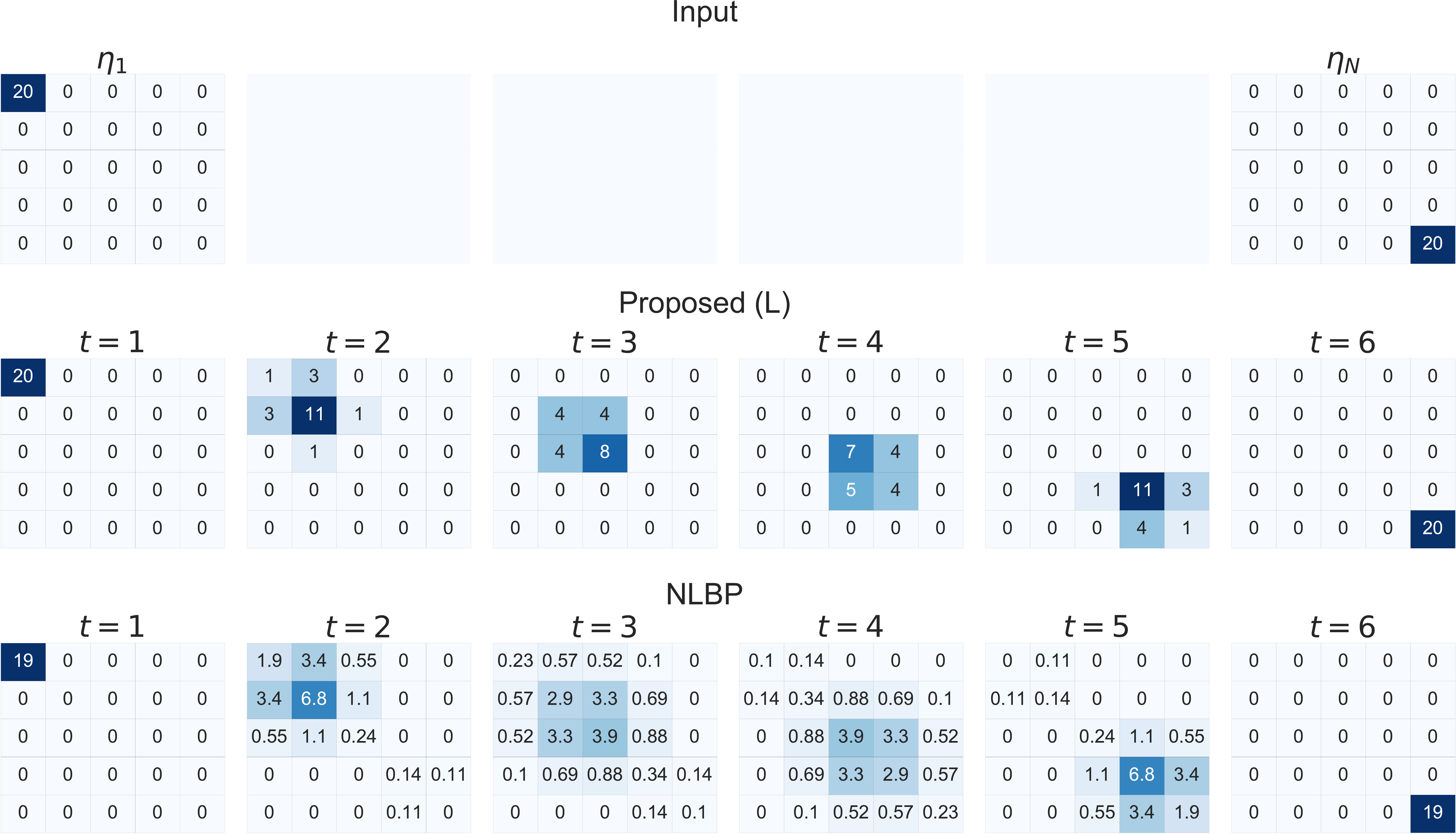}
    }
    \vspace{1.5mm}
    \vspace{1.5mm}
    \fbox{
      \includegraphics[width=1.5\columnwidth]{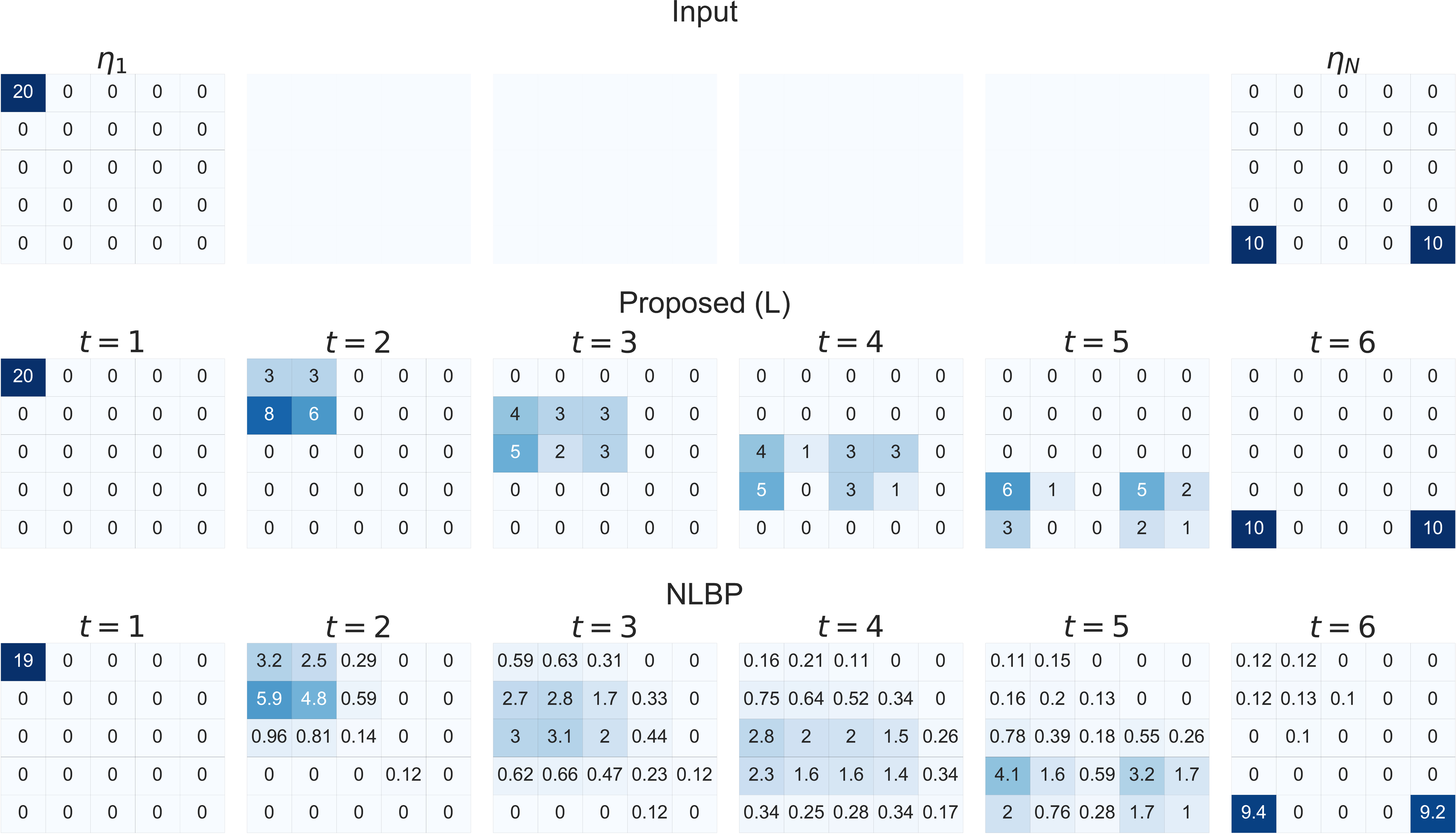}
    }
    \vspace{1.5mm}
    \vspace{1.5mm}
    \fbox{
      \includegraphics[width=1.5\columnwidth]{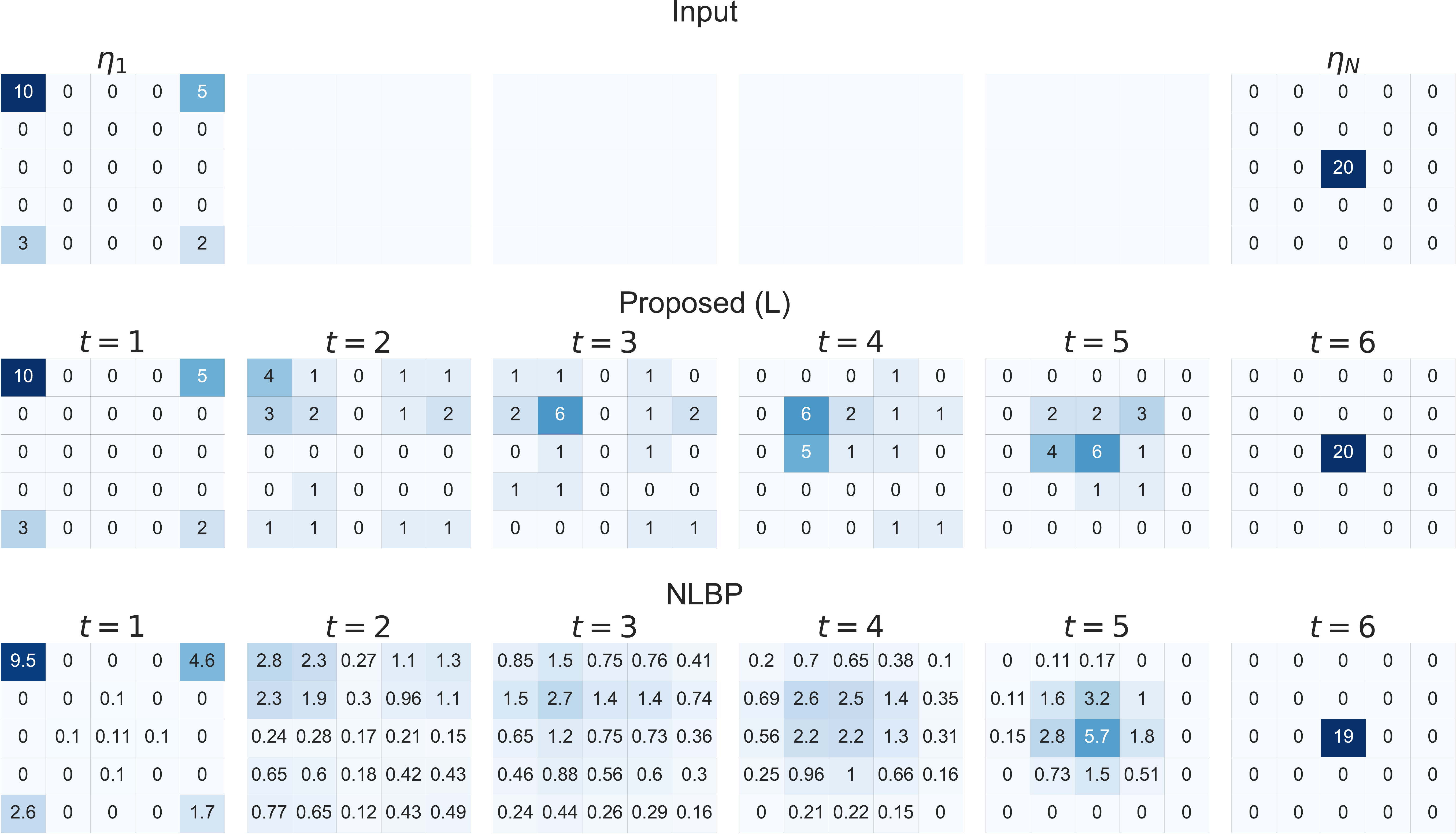}
    }
  \caption{Three examples of interpolation results yielded by each method. In each example, three sequences of histograms in the two-dimensional grid space are presented; the first row shows the input histograms $\bheta_1$ and $\bheta_N$, the second row shows the interpolation results obtained by proposed (L), and the third row shows the interpolation results obtained by NLBP\@. Note that the values smaller than $10^{-2}$ are rounded for visibility.} \label{fig: interpolation} 
\end{figure*}

The results are shown in \cref{fig: interpolation}. 
Note that \cref{fig: interpolation} illustrates different objects from what is shown in \cref{fig: heatmap,fig: heatmap uniform,fig: heatmap distance}; \cref{fig: interpolation} illustrates the interpolated node contingency table values $\tbl_{ti}$ as two-dimensional grid spaces, while \cref{fig: heatmap uniform,fig: heatmap distance} illustrate edge contingency table values $\tbl_{tij}$ as matrices. 
As shown in the figure, NLBP tends to assign non-zero values to many cells, while proposed (L) assigns non-zero values to a small number of cells, resulting in sparse solutions. 
Moreover, the outputs of the proposed (L) are integer-valued while those of NLBP are not. 
This characteristic of the proposed method is beneficial for interpretability when the histogram values are the numbers of countable objects (e.g., the number of people in the area). 

\end{document}